\pdfoutput=1

\documentclass[11pt]{article}

\usepackage{acl}

\usepackage{times}
\usepackage{latexsym}

\usepackage[T1]{fontenc}

\usepackage[utf8]{inputenc}

\usepackage{microtype}

\title{Instructions for *ACL Proceedings}

\usepackage[toc,page,header]{appendix}
\usepackage{titletoc}
\usepackage{caption}
\usepackage{enumitem,kantlipsum}
\usepackage{amsmath}
\usepackage{amsfonts}

\usepackage{amssymb}
\usepackage{subcaption}
\usepackage[utf8]{inputenc} 
\usepackage[T1]{fontenc} 
\usepackage{hyperref} 
\usepackage{url}
\usepackage{booktabs} 
\usepackage{amsfonts} 
\usepackage{nicefrac} 
\usepackage{microtype}
\usepackage{xcolor} 
\usepackage{booktabs}
\def\DETECTOR{\texttt{LAROUSSE}} 
\usepackage{multirow}
\usepackage{multicol}
\usepackage{amssymb}
\usepackage{algpseudocode,algorithm,algorithmicx}
\usepackage{rotating}
\definecolor{lightgray}{gray}{0.55}
\newcommand{\result}[2]{ #1 \color{lightgray}{\scriptsize{$\pm{#2}$}}}
\newtheorem{remark}{Remark}
\usepackage{alphalph}
\def\AUROC{\texttt{AUROC}}
\def\AUPRIN{\texttt{AUPR-IN}}
\def\AUPROUT{\texttt{AUPR-OUT}}
\def\BENCHMARK{\texttt{STAKEOUT}}
\def\FPR{\texttt{FPR}}
\def\ERR{\texttt{Err}}
\def\INPUT{\textbf{INPUT}}
\def\INIT{\textbf{INIT}}
\def\OUTPUT{\textbf{OUTPUT}}

\usepackage{graphicx}

\usepackage{color, colortbl}
\definecolor{LightGreen}{rgb}{1,0.88,1}
\definecolor{LightRed}{rgb}{1,0.8,0.8}

\hypersetup{
    colorlinks=true,
    linkcolor=blue,
    filecolor=magenta,      
    urlcolor=cyan,
    pdftitle={Overleaf Example},
    pdfpagemode=FullScreen,
    }

\title{Toward Stronger Textual Attack Detectors}

\author{Pierre Colombo$^*$$^{2,6}$ \quad Marine Picot$^*$$^{3}$, \\ \textbf{Nathan Noiry}$^{1}$, \textbf{Guillaume Staerman}$^{5}$, \textbf{Pablo Piantanida}$^{4}$ \\ $^1$S2A Telecom Paris, France  \,\ $^2$Equall, Paris, France\,\ \\ $^3$ digeiz, Paris, France \,\ \quad $^4$ ILLS - CNRS, MILA, CentraleSupélec, Canada \\ \,\ $^5$ Universite Paris-Saclay, Inria, CEA,
Palaiseau, France \,\\\ $^6$ MICS, CentraleSupelec, Universite Paris-Saclay, France\\}

\begin{document}
\maketitle

\def\thefootnote{*}\footnotetext{These authors contributed equally to this work}\def\thefootnote{\arabic{footnote}}
\begin{abstract}
The landscape of available textual adversarial attacks keeps growing, posing severe threats and raising concerns regarding the deep NLP system's integrity. However, the crucial problem of defending against malicious attacks has only drawn the attention in the NLP community. The latter is nonetheless instrumental in developing robust and trustworthy systems. This paper makes two important contributions in this line of search: {\it (i)} we introduce {\DETECTOR}, a new framework to detect textual adversarial attacks and {\it (ii)} we introduce {\BENCHMARK}, a new benchmark composed of nine popular attack methods, three datasets, and two pre-trained models. {\DETECTOR} is ready-to-use in production as it is unsupervised, hyperparameter-free, and non-differentiable, protecting it against gradient-based methods. Our new benchmark {\BENCHMARK} allows for a robust evaluation framework: we conduct extensive numerical experiments which demonstrate that {\DETECTOR} outperforms previous methods, and which allows to identify interesting factors of detection rate variations.

\end{abstract}

\section{Introduction}

Despite the high performances of deep learning techniques for Natural Language Processing (NLP) applications, the trained models remain vulnerable to adversarial attacks \cite{barreno2006can, morris-etal-2020-reevaluating} which limits their adoption for critical applications. In the context of NLP, for a given model and a given textual input, an adversarial example is a carefully constructed modification of the initial text such that it is semantically similar to the original text while affecting the model's prediction. The ability to design adversarial examples \cite{alves2018considerations,johnson2018increasing,adarsh_health_ai_2020} raises serious concerns regarding the security of NLP systems. It is, therefore, crucial to develop proper strategies that are available to deal with these threats \cite{Szegedy2014ICLR}. 

\noindent Perhaps surprisingly, if the research community has invested considerable efforts to design efficient attacks, there are only a few works that address the issue of preventing them. One can distinguish two lines of research: {\it detection} methods that aim at discriminating between regular input and attacks; and {\it defense} methods that try to correctly classify adversarial inputs. The latter is based on robust training methods, which customize the learning process, see for instance \cite{zhou2021defense, jones2020robust, yoo-qi-2021-towards-improving,pruthi-etal-2019-combating}. These are limited to certain types of adversarial lures ({\it e.g.}, misspelling), making them vulnerable to other types of attacks that already exist or may be designed in the future. In contrast, detection methods are more relevant to real-life scenarios where practitioners usually prefer to adopt a {\it discard-rather-than-correct} strategy \cite{chow1957optimum}. This has been highlighted in \cite{yoo2022detection} which is, to the best of our knowledge, the single word that introduces a detection method that does not require training. On the contrary, the authors propose to measure the {\it regularity} of a given input by computing the Mahalanobis distance \cite{mahalanobis1936generalized} of its embedding in the last layer of a transformer with respect to the training distribution. Notice that the Mahalanobis distance has also been successfully used in a very similar framework of Out-Of-Distribution (OOD) detection methods (see \cite{mahalanobis,ren2021simple} and references therein).

In this paper, we build upon \cite{yoo2022detection} and introduce a new attack detection framework, called {\DETECTOR}\footnote{{\DETECTOR} stands for textua\underline{\textbf{L}} \underline{\textbf{A}}dversa\underline{\textbf{R}}ial detect\underline{\textbf{O}}r Using half\underline{\textbf{S}}pace ma\underline{\textbf{S}}s d\underline{\textbf{E}}pth}, which improves the current state-of-the-art. Our approach is based on the computation of the {\it halfspace-mass depth} \cite{HM,picot2022adversarial} of the last layer embedding of an input with respect to the training distribution. Halfspace-mass depth is a particular instance of {\it data depth} \cite{darrin2023unsupervised}, which are functions that measure the proximity of a point to the core of a probability distribution. As a matter of fact, the Mahalanobis distance is also -- probably one of the most popular -- a data depth. Interestingly, in addition to improving the attack detection rate, the halfspace-mass depth remedies several limitations of the Mahalanobis depth: it does not make Gaussian assumptions on the data structure and is additionally non-differentiable, providing security guarantees regarding malicious adversaries that could rely on gradient-based methods. 

\noindent The second contribution of our work consists in releasing {\BENCHMARK}, a new NLP attack benchmark that enriches the one introduced in \cite{yoo2022detection}. More precisely, we explore the same datasets and extend their four attacks by adding five new adversarial techniques. This ensures a wider variety of testing methods, leading to a robust evaluation framework that we believe will stimulate future research efforts. We conduct extensive numerical experiments on {\BENCHMARK} and demonstrate the soundness of our {\DETECTOR} detector while studying the main variability factors on its performance. Finally, we empirically observe the presence of relevant information to detect attacks across the layers {\it other} than the last one. This could pave the way for future research by considering the possibility of building detectors that are not limited to the last embedding layers but rather exploit the full network information.

{\bf Our contributions in a nutshell.} Our contributions are threefold:
\begin{enumerate}[wide, labelwidth=!, labelindent=0pt]
    \item We introduce {\DETECTOR}, a {\bf new textual attack detector} based on the computation of a carefully chosen similarity function, the {\it halfspace-mass depth}, between a given input embedding and the training distribution. Contrary to Mahalanobis distance, it does not rely on underlying Gaussian assumptions of data and is non-differentiable, making it robust to gradient-based attacks.
    \item We release {\BENCHMARK}, a {\bf new textual attacks benchmark}, which enriches previous ones by including additional attacks. It contains three datasets and nine attacks, covering a wide range of adversarial techniques, including word/character deletion, swapping, and substitution. This allows for a robust and reliable evaluation framework which will be released in \texttt{DATASETS} \cite{lhoest-etal-2021-datasets} to fuel future research efforts.
    \item We conduct {\bf extensive numerical experiments} to assess the soundness of our {\DETECTOR} detector involving over 20k comparisons, following the method presented in \BENCHMARK. Overall, our results prove that {\DETECTOR} improves the state-of-the-art while being less subject to variability. The code will be released on \url{https://github.com/PierreColombo/AdversarialAttacksNLP}.
\end{enumerate}

The rest of the paper is organized as follows. In \autoref{sec:text_attack_2}, we briefly review the setting of textual attacks, provide main references on the subject, and formally introduce the problem of attack detection. In \autoref{sec:detector_3}, we present our {\DETECTOR} detector and provide some perspectives on data depth and connections to the Mahalanobis distance. In \autoref{sec:benchmark}, we introduce our new benchmark {\BENCHMARK} and give details on the evaluation framework of attack detection. Finally, we present our experimental results in \autoref{sec:expe_results}.

\section{Textual Attacks: Generation and Detection}\label{sec:text_attack_2}

 Let us first introduce some notations. We will denote by $\mathcal{D}= \{ (\mathbf{x}_i, y_i)\}_{1 \leq i \leq n}$ a textual dataset made of $n$ pairs of textual input $\mathbf{x}_i \in \mathcal{X}$ and associated attribute value $y_i \in \mathcal{Y}$. We focus on classification tasks, meaning that $\mathcal{Y}$ is of finite size: $| \mathcal{Y} | < + \infty$. In this work, the inputs are first embedded through a multi-layer encoder with $L$ layers and learnable parameters $\psi \in \Psi$. We denote by $f^\ell_\psi : \mathcal{X} \rightarrow \mathbb{R}^d$ the function that maps the input text to the $\ell$-th layer of the encoder. Note that, as we will work on transformer models, the latent space dimension---the dimension of the output of a layer---of all layers is the same and will be denoted by $d$. The dimension of the logits, denoted as the $(L+1)-$th layer of the encoder, is $d^\prime$. The final classifier built on the pre-trained encoder produces a soft decision $C_\psi$ over the classes, where $\psi$ is a learned parameter. We will denote by $C_\psi(c \, | \, \mathbf{x})$ the predicted probability that a given input $\mathbf{x}$ belongs to class $c$. Given an input $\mathbf{x}$, the predicted label $\hat{y}$ is then obtained as follows:
\begin{align*}
\hat{y} \triangleq \underset{c \in \mathcal{Y}}{\text{arg max}} \; C_{\psi}\left( c| \mathbf{x} \right)
\text{ with }\, C_{\psi} = \text{softmax} (f_{\psi}^{L+1}(\mathbf{x})). 
\end{align*}

\subsection{Review of textual attacks}
The sensitivity of neural networks with respect to adversarial examples has been uncovered by \cite{szegedy2013intriguing} and popularized by \cite{goodfellow2014explaining}, who introduced fast adversarial generation methods, in the context of computer vision. In computer vision, the meaning of an adversarial attack is clear: a given regular input is perturbed by a small noise which does not affect human perception but nonetheless changes the network prediction. However, due to the discrete nature of tokens in NLP, small textual perturbations are usually perceptible (\emph{e.g.}, a word substitution can change the meaning of a sentence). As a result, defining textual attacks is not straightforward and the methods used in the context of images in general do not directly apply to NLP tasks. 

The goal of a textual attack is to modify an input while keeping its semantic meaning and luring a deep learning model. At a high level, one can formally define the problem of textual attack generation as follows. Given an input $\mathbf{x}$, find a perturbation $\mathbf{x}_{adv}$ that satisfies the following optimization problem:
\begin{equation}
 \begin{array}{lr}
       \max \quad  \mathrm{SIM}(\mathbf{x}, \mathbf{x}_{adv}) &  \\
         \mathrm{s.t.} \quad \hspace{0.2cm}  \underset{c \in \mathcal{Y}}{\text{arg max}} \; C_{\psi}\left( c| \mathbf{x}_{adv} \right) \neq \underset{c \in \mathcal{Y}}{\text{arg max}}  \;  C_{\psi}\left( c| \mathbf{x} \right), &
    \end{array}
    \label{eq:adversarial_problem}
\end{equation}
where $\mathrm{SIM}: \mathcal{X} \times \mathcal{X} \rightarrow \mathbb{R}_+$ denotes a function that measures the semantic proximity between two textual inputs. Finding a good similarity function is an active research area and previous works \cite{li2018textbugger} rely on embedding similarities such as Word2vect \cite{mikolov2013efficient}, USE \cite{cer-etal-2018-universal}, or string-based distance \cite{8424632} based on the Levenshtein distance \cite{levenshtein1965leveinshtein}, among others.

The landscape of available adversarial textual attacks keeps growing, with numerous attacks 
every year \cite{li-etal-2021-contextualized,ribeiro-etal-2020-beyond,li-etal-2020-bert-attack,garg-ramakrishnan-2020-bae,alzantot-etal-2018-generating,jia-etal-2019-certified,ren-etal-2019-generating,feng-etal-2018-pathologies,li2018textbugger,zang2019word}. There exist different types of attacks according to the perturbation level, that is the level of granularity at which the corruption is performed. For instance, \cite{ebrahimi-etal-2018-hotflip,pruthi-etal-2019-combating} character-level perturbations are usually based on basic operations such as substitution, deletion, swapping or insertion. There exist also word-level corruption techniques \cite{ebrahimi-etal-2018-hotflip,pruthi-etal-2019-combating} which usually perform word substitution using synonyms or semantically equivalent words \cite{miller1995wordnet,miller1990introduction}. Finally, we can also find sentence-level attacks \cite{iyyer-etal-2018-adversarial} relying on text generation techniques. Standard toolkits such as \texttt{OpenAttack} \cite{zeng-etal-2021-openattack} or \texttt{Textattack} \cite{morris2020textattack} gather them in a unified framework.

\subsection{Review textual attack detection methods}

The goal of an adversarial attack detector is to build a binary decision rule $d : \mathcal{X} \rightarrow \{0,1\}$ that assigns $1$ to \emph{adversarial samples} created by the malicious attacker and $0$ to \emph{clean samples}. Typically, this decision rule consists of a function $s: \mathcal{X} \rightarrow \mathbb{R}$ that measures the similarity between an input sample and the training distribution, and a threshold $\gamma \in \mathbb{R}$: 
\begin{equation}
 d(\mathbf{x}) = \mathbb{I}{\{s(\mathbf{x}) > \gamma\}} = \begin{cases}
1 \quad \text{ if } s(\mathbf{x}) \geq \gamma, \\
0 \quad \text{ if } s(\mathbf{x}) < \gamma .\\
\end{cases} 
\label{eq:detector}
\end{equation}

As already mentioned in the previous section, 
although some works rely on robust training by adding regularization terms that use adversarial generation \cite{dong2021towards,wang2020infobert,yoo-qi-2021-towards-improving} at the risk of not being able to cover attacks developed in the future, adversarial detection techniques have received few attention from the NLP community \cite{mozes-etal-2021-frequency}. Detection methods consist in adding an adversarial attack detector on top of a given trained model. The majority of developed techniques require adversarial examples either for validation or for training purposes. For instance, this is the case of \cite{mozes-etal-2021-frequency} which computes sentence likelihood based on words frequencies; and of \cite{le-etal-2021-sweet,pruthi-etal-2019-combating} which focus on specific types of attacks. The only work that does not require access to adversarial examples is \cite{yoo2022detection} which computes a similarity score between a given input embedding and the training distribution. This similarity function is the Mahalanobis distance and has been widely used in the related literature of OOD detection methods  \cite{podolskiy2021revisiting,ren2021simple,kamoi2020mahalanobis}.

\section{{\DETECTOR}: A Novel Adversarial Attacks  Detector}\label{sec:detector_3}

We follow the notations introduced in \autoref{sec:text_attack_2}. In particular, recall that $f_\psi^L : \mathcal{X} \rightarrow \mathbb{R}^d$ is the mapping to the last layer embedding of the considered network.

\subsection{{\DETECTOR} in a nutshell}

Our framework for adversarial attack detection relies on three consecutive steps: 
\begin{enumerate}[wide, labelwidth=!, labelindent=0pt]

    \item \textbf{Feature Extraction.} As in \cite{yoo2022detection}, we rely on the last layer embedding $f_\psi^L(\mathbf{x})$ of a given textual input $\mathbf{x}$. We will use the following notation: $\mathbf{z} \triangleq f_\psi^L \left( \mathbf{x} \right) \in \mathbb{R}^d$.
    
    \item \textbf{Anomaly Score Computation.} In the second step, we compute a similarity score between the last layer embedding $\mathbf{z}$ and  the predicted class of $\mathbf{z}$. To formally write this score, we need to introduce, for each $y \in \mathcal{Y}$, the empirical distribution $\widehat{P}_Y^L (y)= (1/| \mathcal{D}_y |) \sum_{i: \, y_i = y} \delta_{f_\psi^L(\mathbf{x}_i)}$of the points $\mathcal{D}_y \triangleq \{ f_\psi^L(\mathbf{x}_i), \, \, \mathrm{s.t.} \, y_i = y \}$. With these notations in mind, our similarity score function writes, for a given input $\mathbf{x}$ with predicted class $\hat{y}$:
\begin{equation}
 s_{\DETECTOR}(\mathbf{x}) \triangleq D_{\mathrm{HM}}\big(z, \widehat{P}_Y^L (\hat{y}) \big),
\end{equation}
where $D_{\mathrm{HM}}$ denotes the halfspace-mass depth that we carefully present in \autoref{sec:depth}. The higher the value of $D_{\mathrm{HM}}$ the more regular $\mathbf{x}$ is with respect to $\widehat{P}_{Y}^L$.

    \item \textbf{Thresholding.} Similar to previous works, the final step consists in thresholding our similarity score: we detect $\mathbf{x}$ as an adversarial attack if and only if $s_{\DETECTOR}(\mathbf{x}) \leq \gamma$, where $\gamma$ is a hyperparameter of the detector.
\end{enumerate}

\begin{remark}\label{rq:rq_exp1}
In the experimental section, we will also consider the case where the depth function is computed based on the logits. It corresponds to replace $\mathbf{z} = f_\psi^L \left( \mathbf{x} \right) \in \mathbb{R}^d$ by $\mathbf{z} = f_\psi^{L+1} \left( \mathbf{x} \right) \in \mathbb{R}^{|\mathcal{Y}|}$.
\end{remark}

\subsection{A brief review of data depths and the halfspace-mass depth}\label{sec:depth}

With the goal of extending the notions of order and rank to multivariate spaces,  the statistical concept of depth has been introduced by John Tukey in \cite{Tukey75}. Data depth found many applications in Statistics and Machine Learning (ML) such as in classification \cite{LangeMM14}, clustering \cite{jornsten2004clustering}, text automatic evaluation \cite{dr_distance} or anomaly detection \cite{staerman2020,benchmarkfad}. A depth function $D(\cdot, P): \mathbb{R}^{d}\rightarrow [0,1]$  provides a score that reflects the closeness of any element $\mathbf{x} \in \mathbb{R}^d$ to a probability distribution  $P$ on $\mathbb{R}^d$. The higher (respectively lower) the score of $\mathbf{x}$ is, the deeper (respectively farther) it is in $P$. Many proposals have been suggested in the literature such as the projection depth \cite{Liu92}, the zonoid depth \cite{koshevoy1997} or the Monge-Kantorovich depth  \cite{chernozhukov2017} differing in properties and applications. To compare their benefits and drawbacks, standard properties that a data depth should satisfy have been developed in \cite{ZuoSerfling00} (see also \cite{rainer2004}). We refer the reader to \cite{Mosler13} or to \cite[Ch. 2 ]{phdguigui} for an excellent account of data depth.

{\bf The halfspace-mass depth.} Beyond appealing properties satisfied by depth functions such as affine-invariance \cite{ZuoSerfling00}, these statistical tools suffer in practice from high-computational burden,  which limits their spread use in ML applications \cite{mosler2020choosing}. However, efficient approximations have been provided such as for the halfspace-mass depth \cite{HM} (see also \cite{IRW,AIIRW}). The halfspace-mass (HM) depth of $\mathbf{x}\in \mathbb{R}^d$ w.r.t. a distribution $P$ on $\mathbb{R}^d$ is defined as the expectation over the set of all closed halfspaces containing $\mathbf{x}$ $\mathcal{H}(\mathbf{x})$ of the probability mass of such halfspaces. More precisely,   given a random variable $\mathbf{X}$ following a distribution $P$ and a probability measure $Q$ on $\mathcal{H}(\mathbf{x})$, the HM depth of $\mathbf{x}$ w.r.t. $P$ is defined as follows:
\begin{align}\label{eq:HM}
D_{\mathrm{HM}}(\mathbf{x},P) &= \mathbb{E}_{H\sim Q}\left[ P(H) \right].
\end{align}
 
When a training set $\{\mathbf{x}_1,\ldots, \mathbf{x}_n \}$ is given, expression  \eqref{eq:HM} boils down to: 
 \begin{align}\label{eq:empirique}
    D_{\mathrm{HM}}(\mathbf{x},\hat{P}_X) &= \mathbb{E}_Q\left[ \frac{1}{n}\sum_{i=1}^{n}\mathbb{I} \{\mathbf{x}_i \in H \} \right],  
 \end{align}
where $\hat{P}_X$ denotes the empirical measure defined by $\frac{1}{n}\sum_{i=1}^{n}\delta_{\mathbf{x}_i}$. The halfspace-mass depth has been successfully used in anomaly detection (see \cite{HM} and \cite{AIIRW}) making it a natural candidate for detecting adversarial attacks at the layers of a neural network.


{\bf Computational aspects.} The expectation of \eqref{eq:empirique} can be approximated by means of a  Monte-Carlo as opposed to several depth functions that are defined as the solution to optimization problems \cite{Tukey75,Liu92}, unfeasible when dimensions are too high. The aim is then to approximate \eqref{eq:empirique} with a finite number of half spaces containing $\mathbf{x}$. To that end, authors of \cite{HM} introduced an algorithm, divided into training and testing parts, that provides a computationally efficient approximation of \eqref{eq:empirique}.  The three main parameters involved are $K$, corresponding to the number of directions sampled on the sphere, $n_s$, the sub-sample size which is drawn at each projection step, and $\lambda$, which controls the extent of the choice of the hyperplane. Since the HM approximation has low sensitivity in its parameters, in the remainder of the paper we set $K=10000$, $n_s = 32$ and $\lambda = 0.5$. The computational complexity of the training part is of order $\mathcal{O}(Kn_sd)$ and the testing part $\mathcal{O}(Kd)$, which makes ease to compute. Further details are provided to the curious reader in \autoref{appendix:harcore}
\begin{remark}\textsc{Advantages over the Mahalanobis distance}
In contrast to approaches based on the Mahalanobis distance \cite{mahalanobis,yoo2022detection}, the halfspace-mass depth does not require to invert and estimate the covariance matrix of the training data that can be challenging both from computational and statistical perspectives, especially in high dimension. In addition, the HM depth does not need any assumption on the distribution while Mahanalobis distance is restricted to be used on distributions with finite two-first-order moments.
\end{remark}


\section{{\BENCHMARK}: A Novel Benchmark for Adversarial Attacks }\label{sec:benchmark}

Textual attack generation can be computationally expensive as some attacks require hundreds of queries to corrupt a single sample\footnote{For {\BENCHMARK} the average number of try is 800.}. To dispose of a benchmark that gathers the result of diverse attacks on different datasets and encoders is instrumental to accelerate future research efforts by reducing computational overhead. To build our benchmark, we relied on the models, the datasets, and the attacks available in \texttt{TextAttack} \cite{morris2020textattack}. In the following, we describe the experimental choices we made when building {\BENCHMARK} and discuss our baseline and evaluation pipeline.

\subsection{A novel benchmark: {\BENCHMARK}}

\textbf{Training Datasets.} We choose to work on sentiment analysis, using SST2 \cite{socher-etal-2013-recursive} and IMDB \cite{maas-etal-2011-learning}, and topic classification, relying on ag-news \cite{joachims1996probabilistic}. These datasets are used in \cite{yoo2022detection} and allow for comparison with previously obtained results.

\noindent\textbf{Target Pretrained Classifiers.} We rely on the model available on the Transformers' Hub \cite{wolf-etal-2020-transformers}. In order to ensure that our conclusions are not model specific, we work with classifiers that are based on two types of pre-trained encoders: \texttt{BERT} \cite{devlin-etal-2019-bert} and \texttt{ROBERTA} (\texttt{ROB}) \cite{liu2019roberta}. \autoref{tab:accu} reports the accuracy of the different models on each considered dataset.

\begin{figure}
 \begin{subfigure}[b]{\textwidth}
 \includegraphics[width=0.45\textwidth]{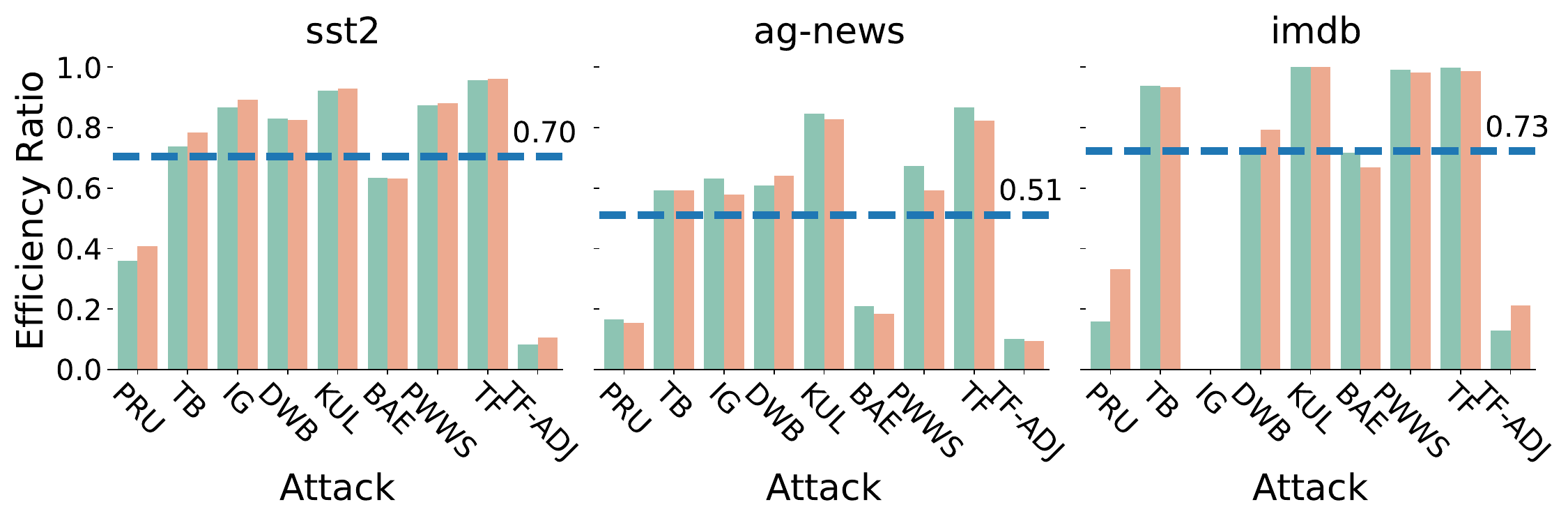}
\end{subfigure}
\begin{subfigure}[b]{\textwidth}
 \includegraphics[width=0.45\textwidth]{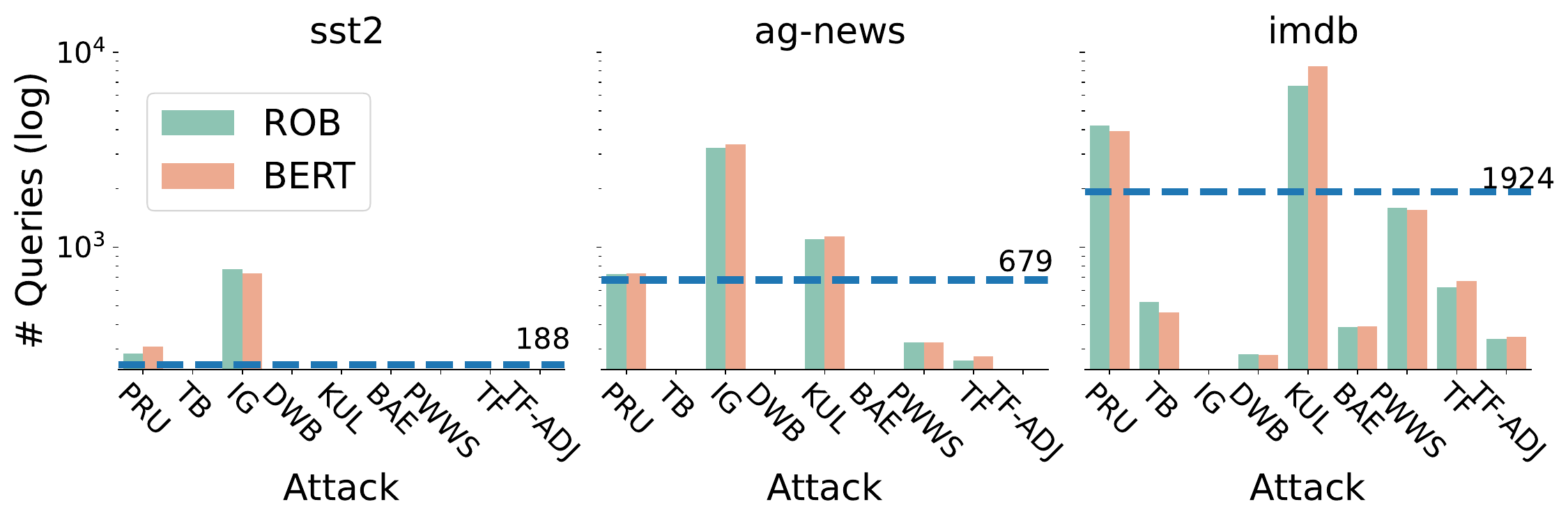}
\end{subfigure}
\caption{Efficiency of the chosen attacks. Both checklist and input reduction were tried but discarded due to low efficiency. Dashed lines report the average performance for each dataset.}
 \label{fig:benchmark_sucess}
 \end{figure}

\begin{table}
\centering
\resizebox{0.25\textwidth}{!}{\begin{tabular}{ccc}\hline
   Model  & Dataset & Acc (\%) \\\hline
  \texttt{BERT}   & \texttt{SST2}& 92.43 \\
      & \texttt{ag-news}& 94.20 \\
       & \texttt{IMDB}&91.90\\
        \texttt{ROB}   &\texttt{SST2} & 94.04\\
      & \texttt{ag-news}&  94.70 \\
       &\texttt{IMDB}& 94.10 \\\hline
\end{tabular}}
 \caption{Classifier accuracy for each considered dataset.}\label{tab:accu} 
\end{table}
\textbf{Adversarial attacks.} Our benchmark is based on 9 different attacks that cover a broad range of techniques including words/character insertion, deletion, swapping, and substitution. Upon these 9 attacks, 8 are taken from the 16 available methods of \texttt{TextAttack}, namely PRUthi (PRU)  \cite{pruthi-etal-2019-combating}, TextBugger (TB) \cite{li2018textbugger}, IGa (IG) \cite{wang2019natural}, DeepWordBug  (DWB) \cite{8424632}, KULeshov (KUL) \cite{kuleshov2018adversarial}, BAE (BAE) \cite{garg-ramakrishnan-2020-bae}, PWW (PWWS) \cite{ren-etal-2019-generating} and TextFooler (TF)\cite{jin2020bert}, and the last one is TF-adjusted (TF-ADF) \cite{morris-etal-2020-reevaluating}. We tried additional attacks, and they were either too weak to fool the models \cite{ribeiro-etal-2020-beyond,feng-etal-2018-pathologies} or were crashing. Further details on the attacks are gathered in \autoref{tab:benchmark_considered_attacks}. \autoref{fig:benchmark_sucess} displays the success rate regarding attack efficiency and the number of queries for each considered attack.  It is worth noting that IG fails on IMDB.

{\bf Takeaways of}  \autoref{fig:benchmark_sucess}. Interestingly, attack efficiency only marginally depends on the pre-trained encoder type. In contrast, there is a strong dependency with respect to the training set (variation of over 0.2 points). It is worth noting that TF and KUL are the most efficient attacks. From the averaged number of queries, we note that attacking a classifier trained on IMDB is harder than one trained on SST2 despite being both binary classification tasks.
\\\noindent\textbf{Adversarial and clean sample selection.} For evaluation, we rely on test sets that are made of clean samples and adversarial ones. In order to construct such sets while controlling the ratio between clean and adversarial samples, we rely on \cite[Scenario 1]{yoo2022detection}. From a given initial test set $\mathcal{X}_t$, we sample two disjoint subsets $\mathcal{X}_1$ and $\mathcal{X}_2$. We then generate attacks on $\mathcal{X}_1$ and take the successful one as an adversarial testing example, while $\mathcal{X}_2$ is taken as the clean testing sample.

\subsection{Baseline detectors} 
We use two baseline detectors. The first one is based on a language model likelihood and the second one corresponds to the Mahalanobis detector introduced in \cite{yoo2022detection}. Both of them follow the analog three consecutive steps of {\DETECTOR}, but do not use the same similarity score.

{\bf Language model score.} This method consists in computing the likelihood of an input with an external language model:
\begin{equation}\label{eq:gpt_score}
 s_{\mathrm{LM}}(\mathbf{x}) = - \sum_{i=1}^{|\mathbf{x}|} \log p_{\psi}(\omega_i| \omega_{i-1}, \ldots, \omega_1),
 \end{equation}
 where $\omega_i$ represents the individual token of the input sentence $\mathbf{x}$.
We compute the log-probabilities with the output of a pretrained \texttt{GPT2} \cite{NEURIPS2020_1457c0d6}. Notice that this baseline is also used in \cite{yoo2022detection}.

{\bf Mahalanobis-based detector.} We follow \cite{yoo2022detection} which relies on a class-conditioned Mahalanobis distance. Following our notations, it corresponds to evaluation: 
\begin{equation}\label{eq:maha_score}
s_{\mathrm{M}}(\mathbf{x}) = \left(f_\psi^L(\mathbf{x}) - \mu_{\hat{y}}\right)^T \Sigma_{\hat{y}}  \left(f_\psi^L(\mathbf{x}) - \mu_{\hat{y}}\right),
\end{equation}
where $\mu_{\hat{y}}$ is the empirical mean for the logits of class $\hat{y}$ and $\Sigma_{\hat{y}}$ is the associated empirical covariance.

\begin{remark}
Similarly to Remark \autoref{rq:rq_exp1}, for a given textual input $\mathbf{x}$, we will either rely on the penultimate layer $L$ representation $f_\psi^L(\mathbf{x})$ or on the logits predictions $f_\psi^{L+1}(\mathbf{x})$ of the networks to compute $s_M$.
\end{remark}


\begin{table*}
 \centering
 \caption{Aggregated performance over both datasets and attacks. Each average number aggregates 270 measurements (10 seeds $\times$ 3 datasets $\times$ 9 attacks). $D_M$ (resp. $D_{HM}$) indicates a detector based on Mahalanobis (resp. Halspace Mass depth) (see \autoref{eq:maha_score}), GPT to the perplexity score (see \autoref{eq:gpt_score}).} {Purple color corresponds to {\DETECTOR}}.\label{tab:all_average_main}
\resizebox{0.75\textwidth}{!}{\begin{tabular}{llllllll}\toprule
&&&{\AUROC} &{\FPR} &{\AUPRIN} &{\AUPROUT}&{\ERR} \\
\midrule\texttt{BERT}& $GPT$ & softmax&\result{76.1} {9.1} & \result{ 58.4} {19.1} &\result{75.4 }{8.7} & \result{75.3} {10.2 }&\result{34.0} {9.5} \\
 & $D_{\mathrm{M}}$ & $f_\psi^L$ &\result{88.8} {6.3} & \result{ 49.7} {25.4} &\result{90.9 }{6.0} & \result{84.5} { 8.2 }&\result{28.4} { 12.0} \\
&& $f_\psi^{L+1}$ &\result{90.1} {7.8} & \result{ 32.3} {23.9} &\result{88.6 }{ 10.4} & \result{88.5} { 7.6 }&\result{19.7} { 11.3} \\ 
\rowcolor{LightGreen} & $D_{\mathrm{HM}}$ & $f_\psi^L$&\textbf{\result{92.0} {5.0}} & \textbf{\result{ 32.1} {24.1}} &\textbf{\result{93.3 }{4.8}} &\textbf{ \result{89.4} { 5.8 }}&\textbf{\result{19.5} { 11.2}} \\ 
&& $f_\psi^{L+1}$ &\result{91.9} {5.1} & \result{ 35.8} {23.2} &\result{92.4 }{5.7} & \result{90.0} { 5.6 }&\result{21.4} { 10.9} \\ \hline
\texttt{ROB.} & $GPT$ & softmax&\result{77.7} {9.7} & \result{ 56.0} {20.4} &\result{77.2 }{9.1} & \result{76.8} {10.7 }&\result{32.6} {9.9} \\ 
 & $D_{\mathrm{M}}$ & $f_\psi^L$&\result{89.9} {5.5} & \result{ 44.1} {22.9} &\result{91.9 }{5.1} & \result{86.1} { 7.2 }&\result{25.5} { 10.9} \\
&& $f_\psi^{L+1}$ &\result{90.0} {8.3} & \result{ 31.9} {23.6} &\result{88.5 }{ 11.5} & \result{88.7} { 7.8 }&\result{19.5} { 11.3} \\ 
\rowcolor{LightGreen} & $D_{\mathrm{HM}}$ & $f_\psi^L$&\textbf{\result{93.4} {4.6}} & \textbf{\result{ 29.0} {21.7}} &\textbf{\result{93.9 }{5.4}} & \textbf{\result{91.3} { 5.3 }}&\textbf{\result{17.9} { 10.3}} \\
&& $f_\psi^{L+1}$ &\result{92.8} {5.1} & \result{ 32.1} {23.5} &\result{93.3 }{5.9} & \result{90.9} { 5.9 }&\result{19.4} { 11.3} \\
\bottomrule
\end{tabular} }
\end{table*}
\subsection{Evaluation metrics}
The adversarial attack detection problem can be seen as a classification problem. In our context, two quantities are of interest, namely {\it (i)} the {\it false alarm rate}, \textit{i.e.} the proportion of samples that are misclassified as \emph{adversarial sample} while actually being \emph{clean};  and {\it (ii)} the {\it true detection rate}, \textit{i.e.,} the proportion of samples that are rightfully predicted as \emph{adversarial sample}.
We focus on three different metrics that assess the quality of our method. 
\begin{enumerate}[wide, labelwidth=!, labelindent=0pt]
    \item \textbf{Area Under the Receiver Operating Characteristic curve  ({\AUROC};  \cite{bradley1997use}).} It is the area under the ROC curve which consider the true detection rate against the false alarm rate. From elementary computations, the {\AUROC} can be linked to the probability that a clean example has higher score than an adversarial sample.
  \item \textbf{Area Under the Precision-Recall curve (\texttt{AUPR}; \cite{davis2006relationship})}. It is the area under the precision-recall curve that is more relevant to imbalanced situations. It plots the recall (true detection rate) against the precision (actual proportion of \emph{adversarial sample} amongst the predicted \emph{adversarial sample}).
 \item \textbf{False Positive Rate at 90\% True Positive Rate ({\FPR} (\%))}. In a practical situation, one wishes to build an efficient detector. Thus, given a detection rate $r$, this incites to fix a threshold $\delta_r$ such that the corresponding TPR equals $r$. Following \cite{yoo2022detection}, we set $r = 0.90$. For {\FPR}, lower is better.
 \item \textbf{Classification error ({\ERR} (\%))}. This refers to the lowest classification error obtained by choosing the best-fixed threshold.
\end{enumerate}

\section{Experimental results}\label{sec:expe_results}

\subsection{Overall Results}\label{ssec:overall_results}
We report in \autoref{tab:all_average_main} the aggregated performance over the different datasets, the various seeds, and the different attacks.
\\\noindent \textbf{$D_{\mathrm{HM}}$ achieves the best overall results.} It is worth noting that detection methods better discriminate adversarial attacks on \texttt{ROB.} than \texttt{BERT}. It also consistently improves the performance when using a halfspace mass score $D_{\mathrm{HM}}$ instead of Mahanalobis $D_{\mathrm{M}}$, which experimentally validates our choice. This conclusion holds on both \texttt{ROB.} and \texttt{BERT}, corresponding to over $540$ experimental configurations. Similar to previous work \cite{yoo2022detection}, the detector built on \texttt{GPT2} under-performs $D_{\mathrm{M}}$. For all methods, we observe that {\DETECTOR} achieves the best results both in terms of threshold-free (\textit{e.g.} {\AUROC}, {\AUPRIN} and {\AUPROUT}) threshold-based metrics (\textit{e.g.} {\FPR}) which validates our detector.
\begin{figure}
\centering
\begin{subfigure}[b]{0.48\textwidth}
 \centering
 \includegraphics[width=\textwidth]{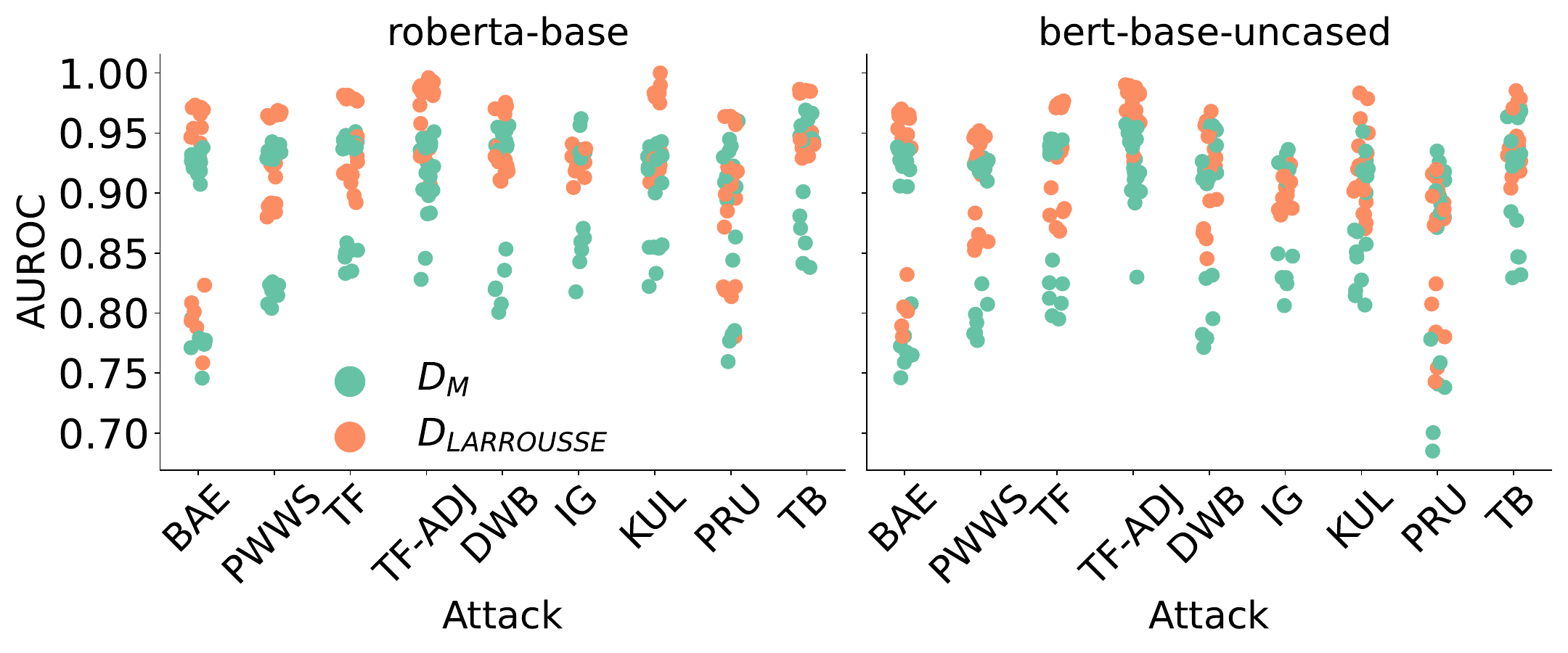}
\end{subfigure}
\begin{subfigure}[b]{0.48\textwidth}
 \centering
 \includegraphics[width=\textwidth]{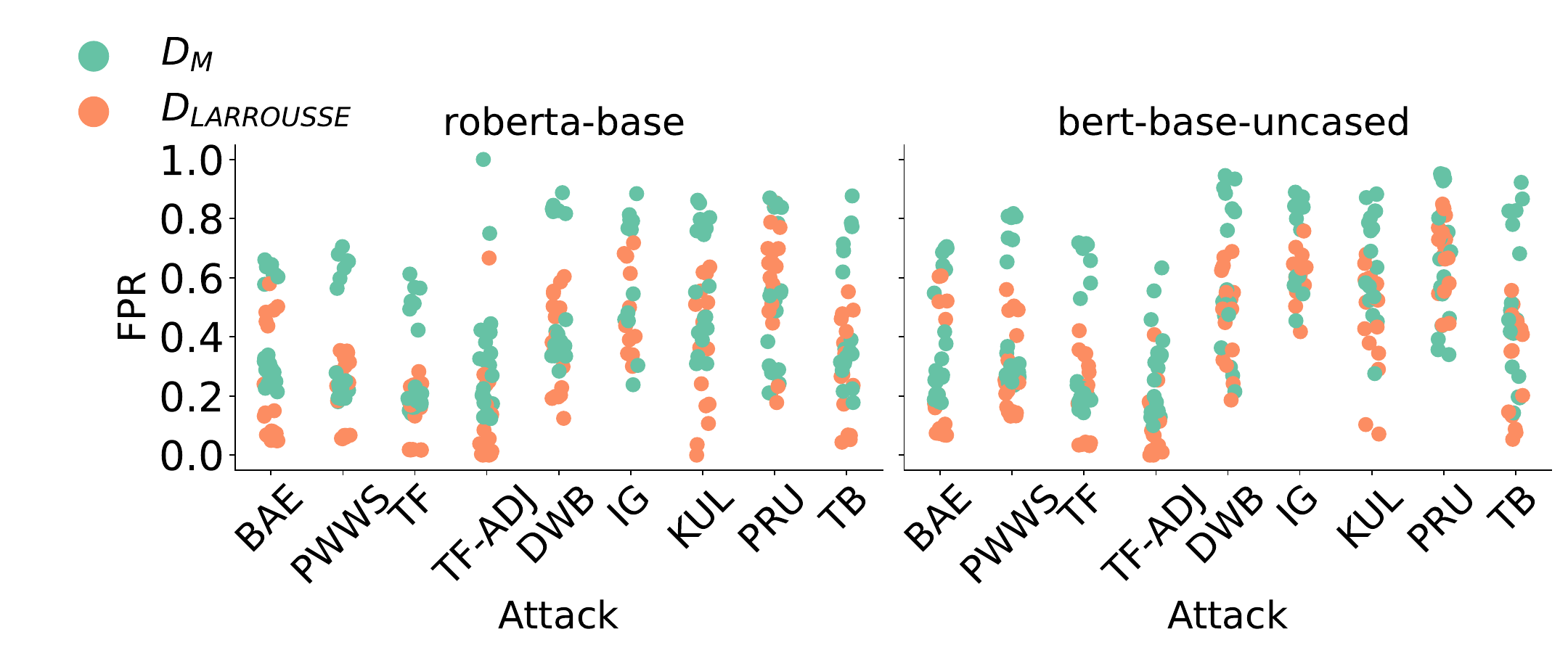}
\end{subfigure}
\caption{Performance per attack in for each pretrained encoder in terms of {\AUROC} (left) and {\FPR} (right)  of $D_{\mathrm{M}}$ and $D_{\mathrm{HM}}$ on {\BENCHMARK}. }\label{fig:per_attack_per_model} \vspace{-0.3cm}
\end{figure}
\\\noindent \textbf{Importance of feature selection for adversarial detectors.} Both $D_{\mathrm{HM}}$ and $D_{\mathrm{M}}$ are highly sensitive to the layer's choice. For $D_{\mathrm{M}}$, using the logits is better than the penultimate layer, while for $D_{\mathrm{HM}}$, the converse works better. Although the {\AUROC} presents a slight variation when using $f_\psi^{L+1}$ instead of $f_\psi^{L}$, it induces a variation of over 10 {\FPR} points.

Overall, it is worth noting that {\DETECTOR}, although being state-of-the-art on tested configurations, achieves an {\FPR} which remains moderate. The best-averaged error of $17.9\%$ is far from the error achieved on the main task (less than 10\% on all datasets).

\begin{figure*}[h]
\centering
\begin{subfigure}[b]{0.29\textwidth}
 \centering
 \includegraphics[width=\textwidth]{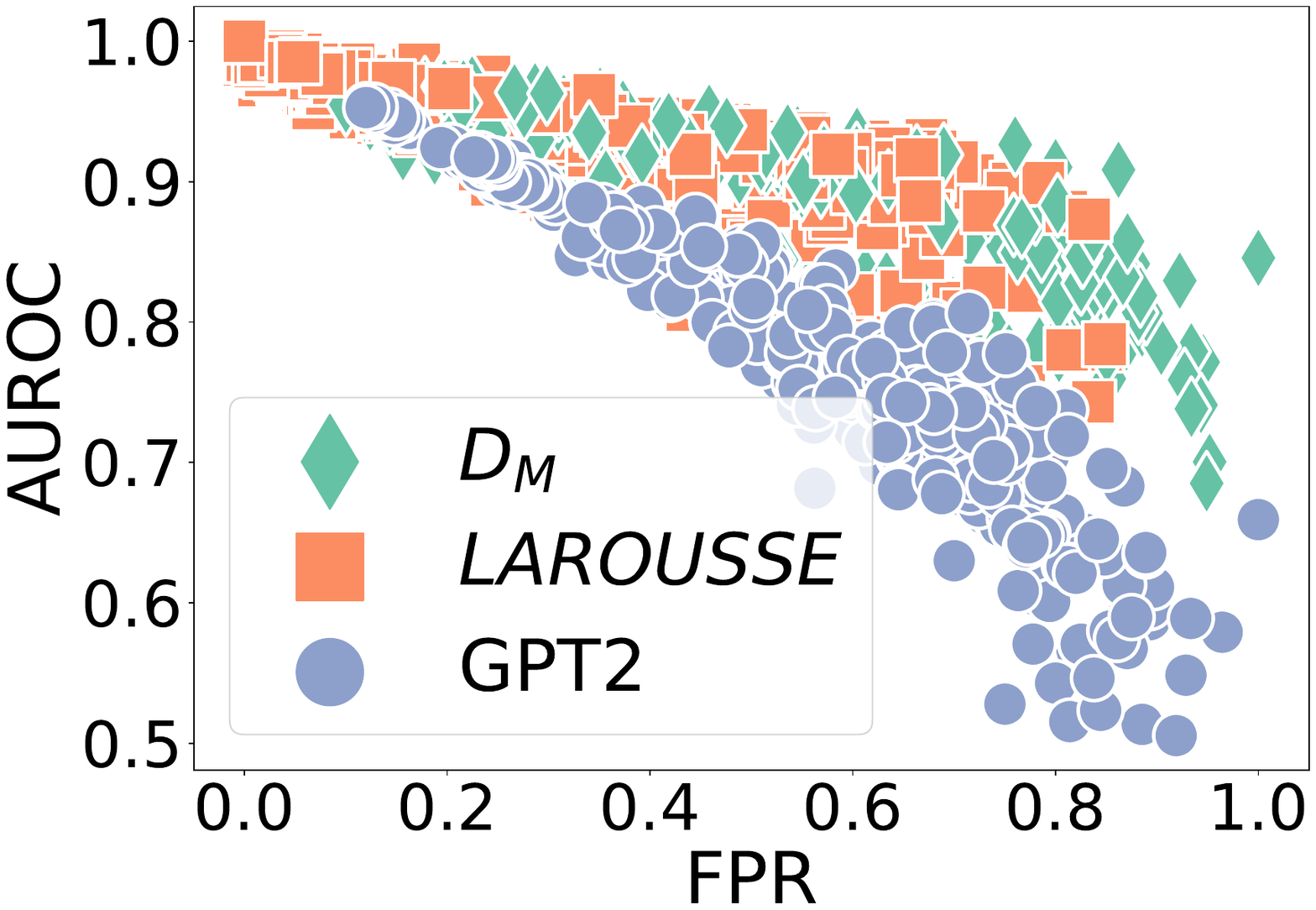}
\end{subfigure}
\begin{subfigure}[b]{0.29\textwidth}
 \centering
 \includegraphics[width=\textwidth]{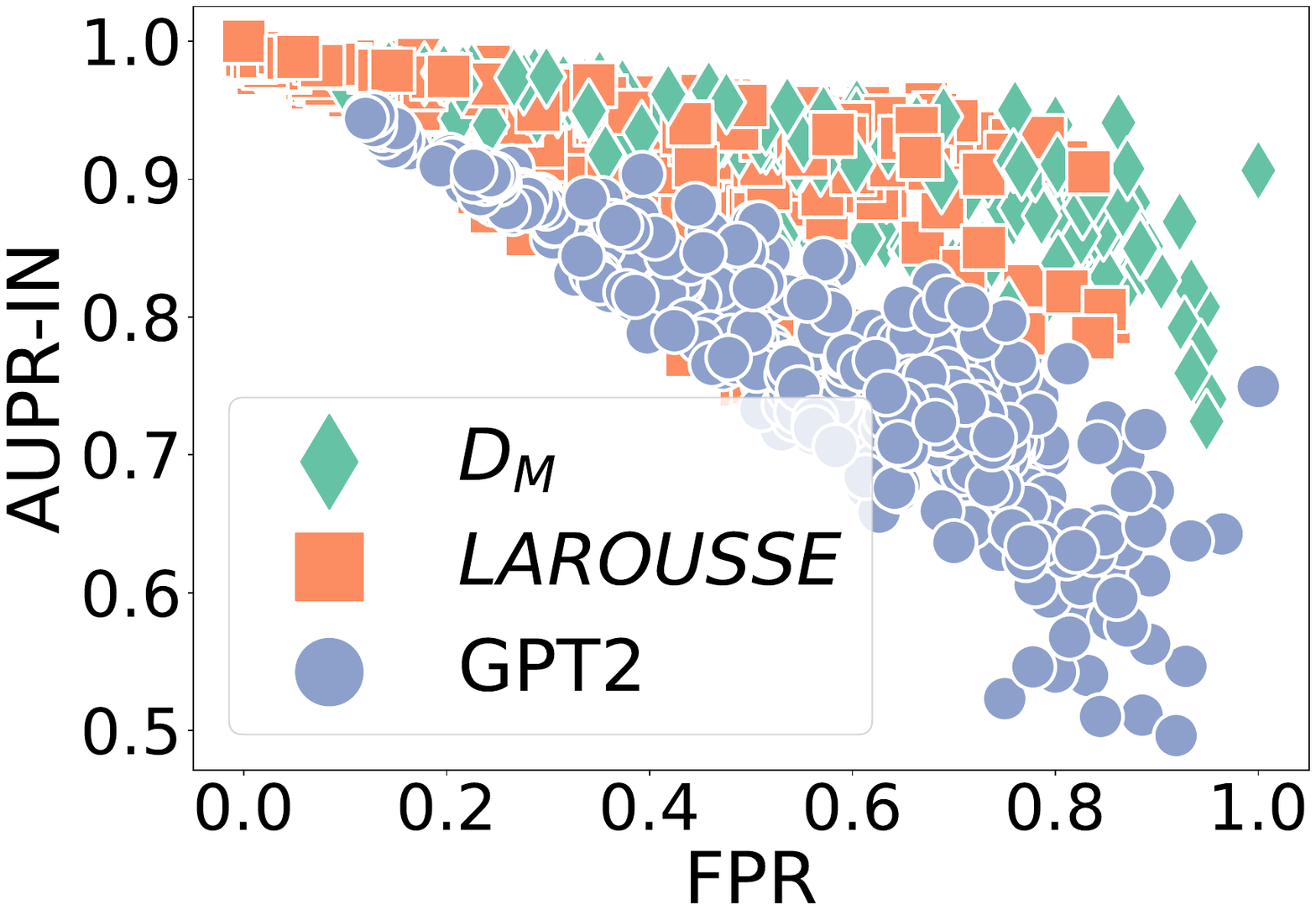}
\end{subfigure}
\begin{subfigure}[b]{0.29\textwidth}
 \centering
 \includegraphics[width=\textwidth]{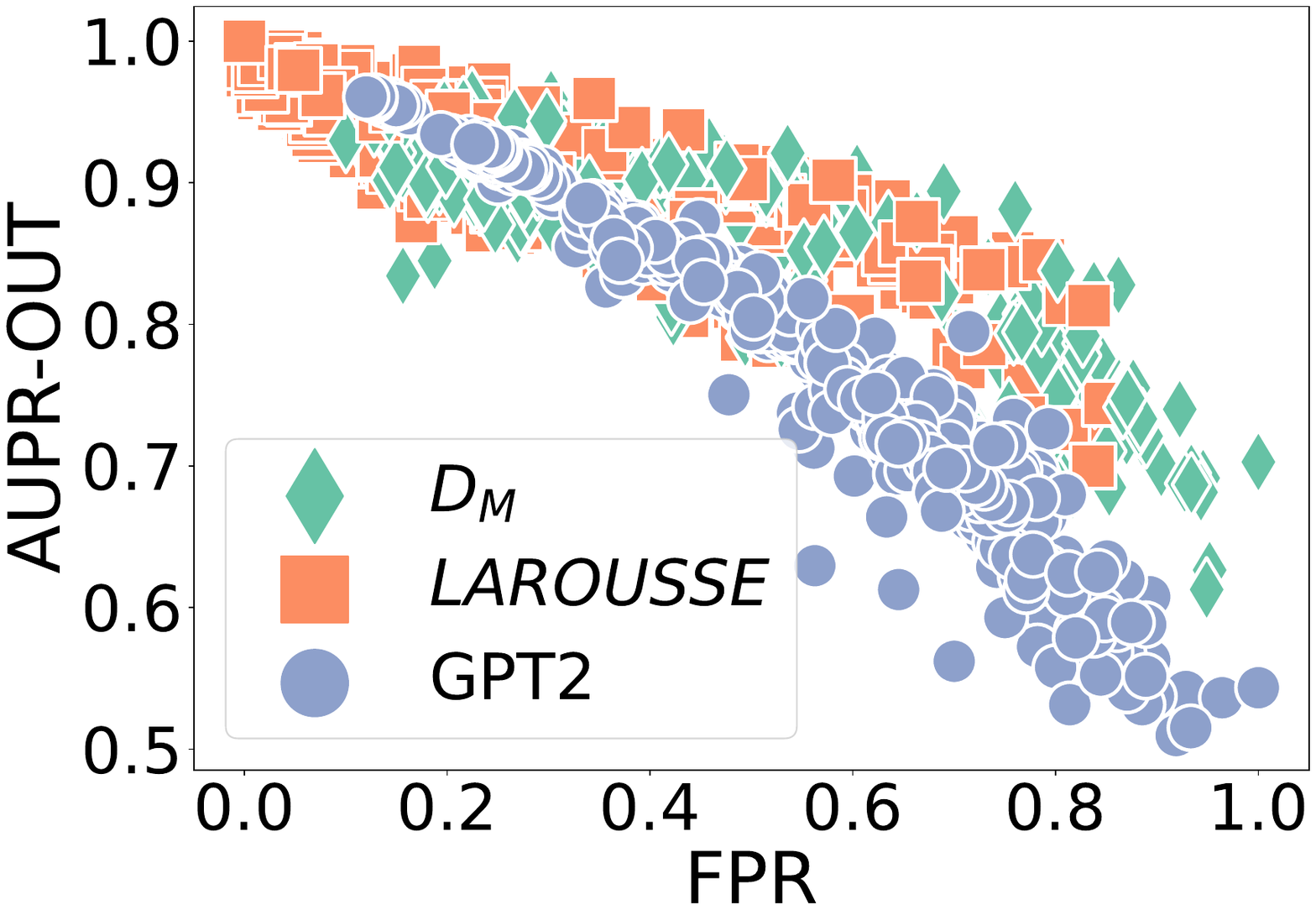}
\end{subfigure}
\caption{Empirical study of the metric relationship for the three considered detection methods.}
\label{fig:all_tradeoff_main}
\end{figure*}

\subsection{Identifying key detection factors}\label{ssec:identifying_key}
\begin{figure}

\begin{subfigure}[b]{\textwidth}

 \includegraphics[width=0.45\textwidth]{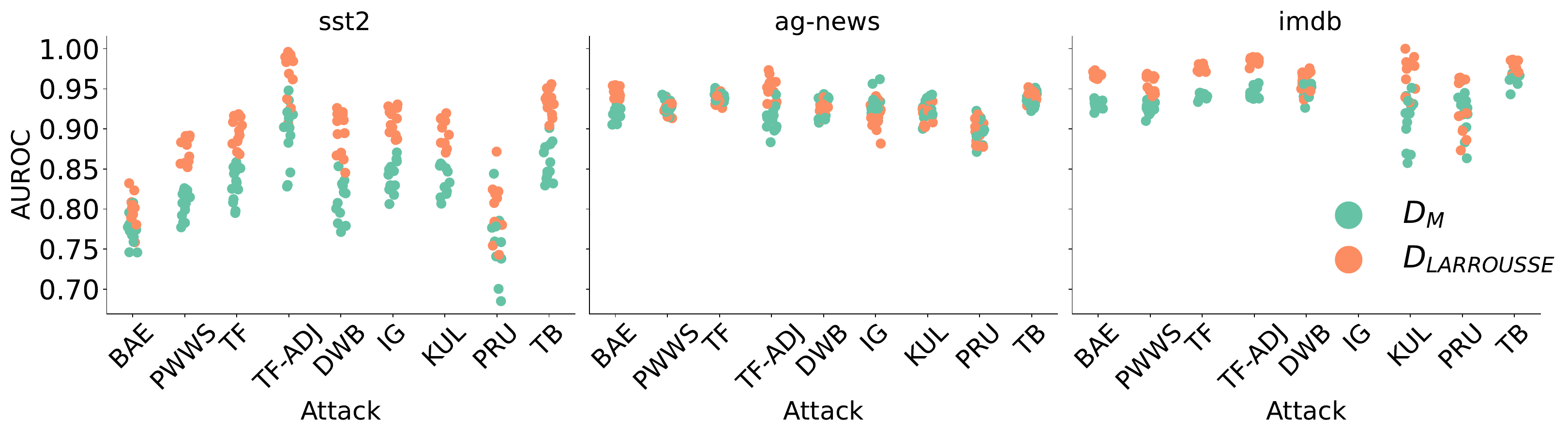}
\end{subfigure}
\begin{subfigure}[b]{\textwidth}

 \includegraphics[width=0.45\textwidth]{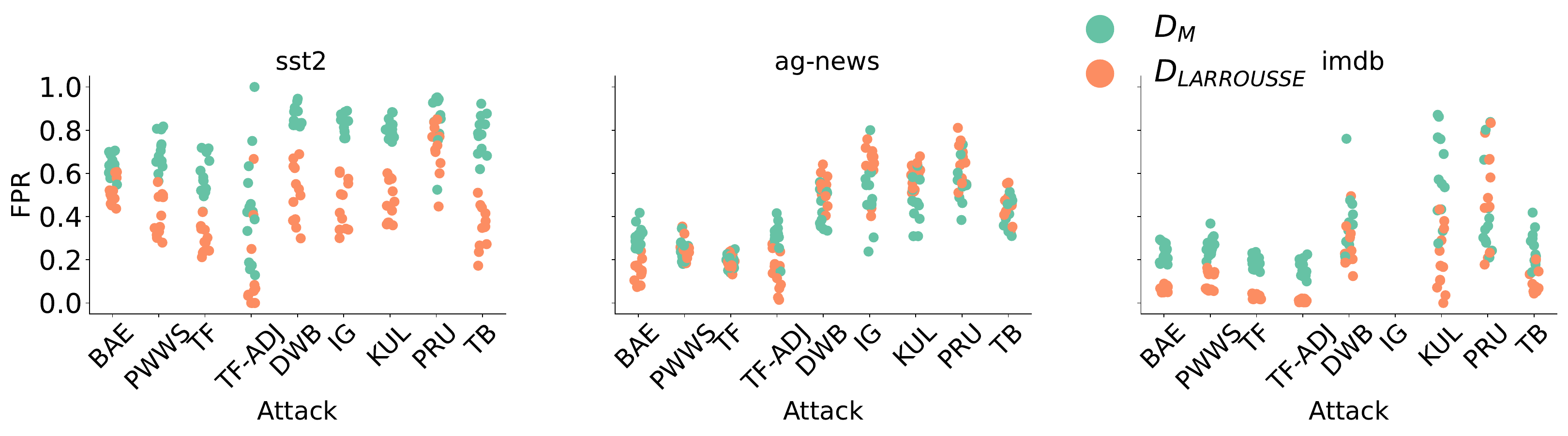}
\end{subfigure}
\caption{Performance in terms of {\AUROC} (up) and {\FPR} (down)  of $D_{\mathrm{M}}$ and $D_{\mathrm{HM}}$ on {\BENCHMARK}. \autoref{fig:all_results_per_attack_appendix} in the Supplementary Material reports the results of \texttt{GPT2}.} 
\label{fig:all_plots_per_model} \vspace{-0.3cm}
\end{figure}

To better understand the performance of our methods w.r.t different attacks and various datasets, we report in \autoref{fig:all_plots_per_model} the performance in terms of {\AUROC} and {\FPR} per attack. 
\\\noindent \textbf{Detectors and models are not robust to dataset change.} The detection task is more challenging for SST-2 than for ag-news and IMDB, with a significant drop in performance  (\textit{e.g.} over 15 absolute points for BAE). On SST-2, $D_{\mathrm{HM}}$ achieves a significant gain over $D_{\mathrm{M}}$ both for the {\AUROC} and {\FPR}. \\\noindent \textbf{Detectors do not detect uniformly well the various attacks.} This phenomenon is pronounced on SST2 while being present for both ag-news and IMDB. For example on SST2, {\FPR} varies from less than 10 (a strong detection performance) for TF-ADJ to over 70 (a poor performance) for PRU. 
\\\noindent \textbf{Hard to detect attack for \texttt{ROB.} are not necessarily hard to detect for \texttt{BERT}.} This phenomenon is illustrated by \autoref{fig:per_attack_per_model}. For example, KUL is hard to detect for \texttt{BERT} while being easier on \texttt{ROB} as {\DETECTOR} achieves over 96 {\AUROC} points. If safety is a primary concern, it is thus crucial to carefully select the pre-trained encoder.
\\\noindent \textbf{The choice of clean samples largely affects the detection performance measure.} \autoref{fig:all_plots_per_model} and \autoref{fig:per_attack_per_model} display several tries with different seeds. As mentioned in \autoref{sec:benchmark}, different seeds correspond to various choices of clean samples. On all datasets, we observe that when measuring the algorithm performance, different negative samples will lead to different results (\textit{e.g.} {\FPR} on IMDB varies of over 30 points on KUL and PRU across different seeds).

\subsection{All the metrics matter}
\textbf{Setting.} In this experiment, we study the relationship between the different metrics. From \autoref{tab:all_average_main}, we see that threshold free metrics (\textit{i.e.}, {\AUROC}, {\AUPRIN}) exhibit lower variance than threshold based metrics such as {\FPR}. The {\FPR} measures the percentage of natural samples detected as adversarial when $90\%$ of the attacked samples are detected. Therefore, the lower, the better. 
\\\noindent \textbf{Takeaways.} From \autoref{fig:all_tradeoff_main}, we see that for a large {\AUROC}, {\AUPRIN} and {\AUPROUT} do not necessarily corresponds a low {\FPR}. This suggests that the detectors also detect natural samples as adversarial when it detect at least 90$\%$ of adversarial examples. Additionally, a small variation of {\AUROC}, {\AUPRIN} and {\AUPROUT} can lead to a high change in {\FPR}. It is therefore crucial to compare the detectors using all metrics. 
\subsection{Expected performances of {\DETECTOR}}
\begin{figure}
\centering
    \includegraphics[width=0.35\textwidth]{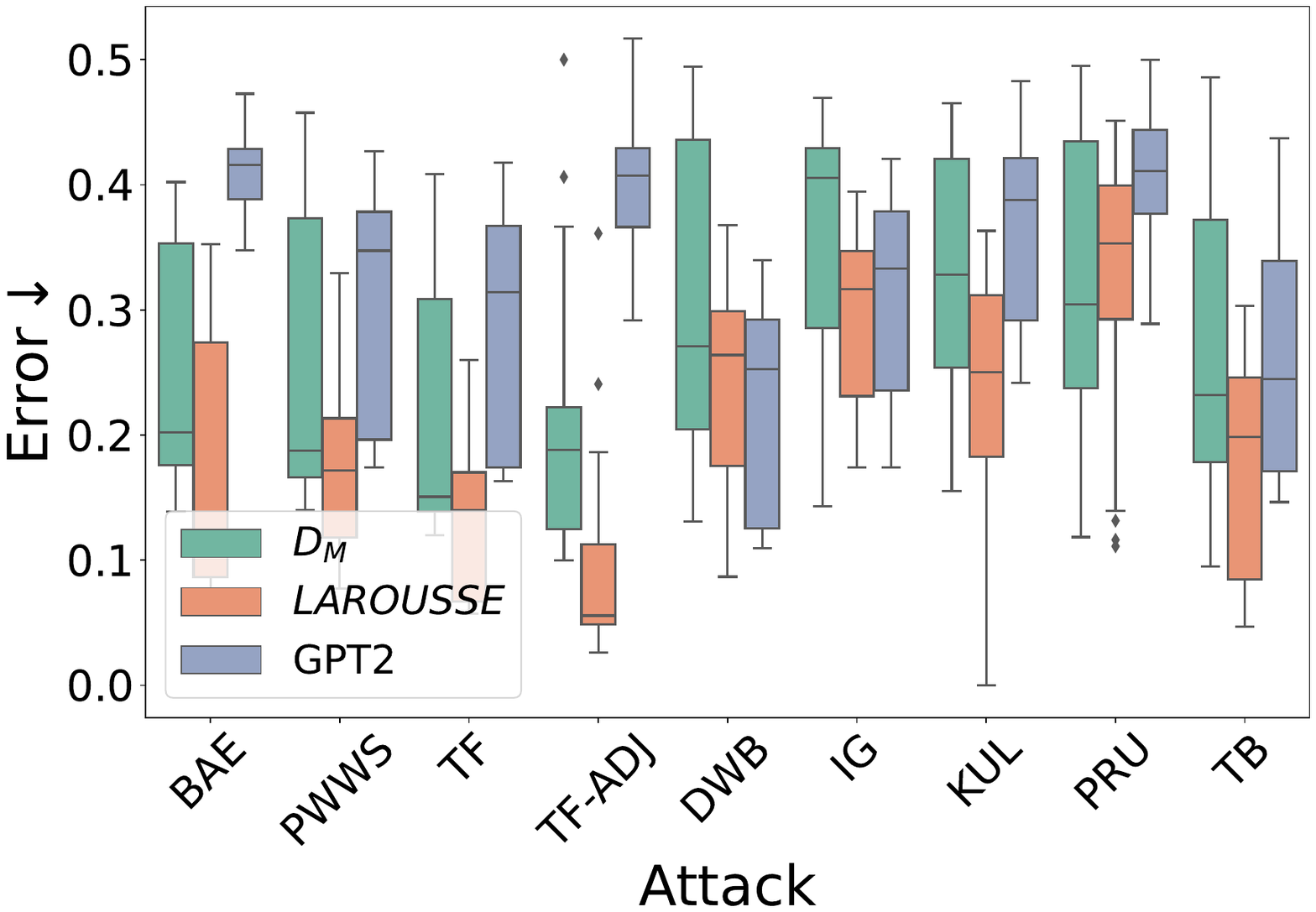}
    \caption{Detection error on {\BENCHMARK}.}\label{fig:crafting_a_detector}\vspace{-0.3cm}
\end{figure}

\textbf{Setting.} \autoref{fig:crafting_a_detector} reports the error probability per attack for {\DETECTOR} and the considered baselines.

\noindent \textbf{Efficient attacks are easier to detect.} We observe that on the three most efficient attacks, according to  \autoref{fig:benchmark_sucess} (\textit{i.e.,} TF, PWWS and KUL), {\DETECTOR} is significantly more effective than $D_{\mathrm{M}}$ and \texttt{GPT2}.
\\\noindent \textbf{Different detection methods capturing various phenomena are better suited for detecting types of attack.}  Although {\DETECTOR} achieves the best results overall, \texttt{GPT2}, which relies on perplexity solely, achieves competitive results with {\DETECTOR} and outperforms $D_{\mathrm{M}}$ on several attacks (\textit{i.e.,} DWB and IG). This suggests that stronger detectors could be achieve by combining different types of scoring functions.

 \subsection{Semantic vs syntactic attacks}
In this section, we analyze the results of the {\DETECTOR} on semantic (\textit{i.e.,} working on token) versus syntactic (\textit{i.e,} working on character) attacks. Raw and processed results are reported in \autoref{ssec:syntactic}.

\textbf{Takeaways.} From \autoref{fig:all_fpr_sem}, we observe that semantic attacks are harder to detect for both our method and $D_H$.

 \begin{figure}[!ht]
\centering
 \includegraphics[width=0.3\textwidth]{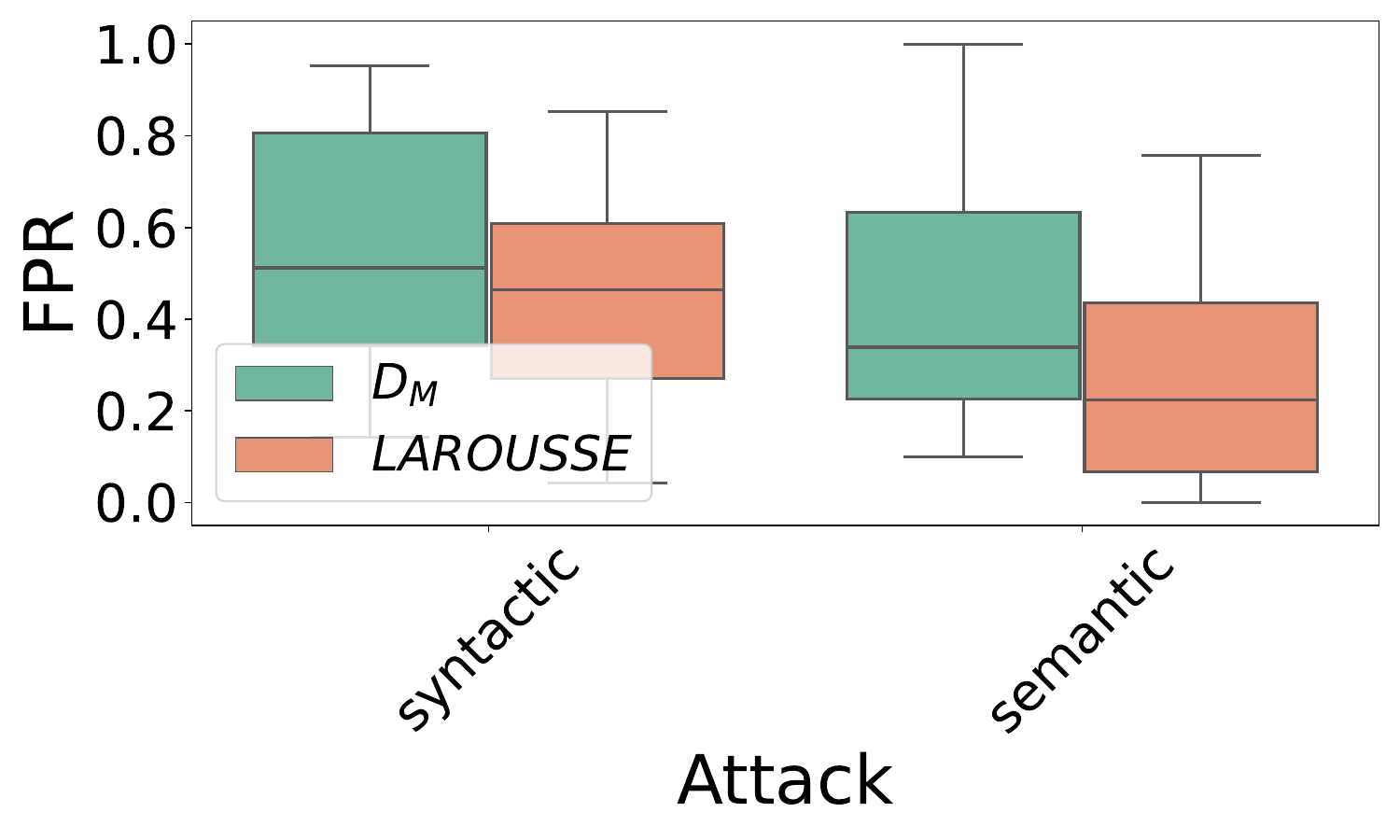}
 \caption{{\FPR} for semantic vs syntactic analysis further results can be found in \autoref{fig:y equals x}} 
 \label{fig:partical}
\end{figure}

\section{Concluding Remarks}

\vspace{-0.3cm}

We have proposed {\BENCHMARK}, a large adversarial attack detection benchmark, and {\DETECTOR}. {\DETECTOR} leverages \emph{a new anomaly score built on the halfspace-mass depth} and offers a better alternative than the widely known Mahalanobis distance. 

\section{Ethical impact of our work}
Our work focuses on responsive NLP and aims at contributing to the protection of NLP systems. Our new benchmark {\BENCHMARK} allows for a robust evaluation of new adversarial detection methods. And {\DETECTOR} outperforms previous methods and thus provides a better defense against attackers. Overall, we believe this paper offers promising research direction toward safe and robust NLP systems and will benefit the community. 

\section*{Acknowledgements}
This work was performed using HPC resources from GENCI-IDRIS (Grants 2022- AD01101838, 2023-103256 and 2023-101838).



\clearpage

\bibliographystyle{plainnat}
\bibliography{bib}

\begin{thebibliography}{103}
\providecommand{\natexlab}[1]{#1}
\providecommand{\url}[1]{\texttt{#1}}
\expandafter\ifx\csname urlstyle\endcsname\relax
  \providecommand{\doi}[1]{doi: #1}\else
  \providecommand{\doi}{doi: \begingroup \urlstyle{rm}\Url}\fi

\bibitem[Aldahdooh et~al.(2022{\natexlab{a}})Aldahdooh, Hamidouche, and
  D{\'e}forges]{aldahdooh2022revisiting}
Ahmed Aldahdooh, Wassim Hamidouche, and Olivier D{\'e}forges.
\newblock Revisiting model’s uncertainty and confidences for adversarial
  example detection.
\newblock \emph{Applied Intelligence}, pages 1--23, 2022{\natexlab{a}}.

\bibitem[Aldahdooh et~al.(2022{\natexlab{b}})Aldahdooh, Hamidouche, Fezza, and
  D{\'e}forges]{aldahdooh2022adversarial}
Ahmed Aldahdooh, Wassim Hamidouche, Sid~Ahmed Fezza, and Olivier D{\'e}forges.
\newblock Adversarial example detection for dnn models: A review and
  experimental comparison.
\newblock \emph{Artificial Intelligence Review}, pages 1--60,
  2022{\natexlab{b}}.

\bibitem[Alves et~al.(2018)Alves, Bhatt, Hall, Driscoll, Murugesan, and
  Rushby]{alves2018considerations}
Erin Alves, Devesh Bhatt, Brendan Hall, Kevin Driscoll, Anitha Murugesan, and
  John Rushby.
\newblock Considerations in assuring safety of increasingly autonomous systems.
\newblock \emph{NASA}, 2018.

\bibitem[Alzantot et~al.(2018)Alzantot, Sharma, Elgohary, Ho, Srivastava, and
  Chang]{alzantot-etal-2018-generating}
Moustafa Alzantot, Yash Sharma, Ahmed Elgohary, Bo-Jhang Ho, Mani Srivastava,
  and Kai-Wei Chang.
\newblock Generating natural language adversarial examples.
\newblock In \emph{Proceedings of the 2018 Conference on Empirical Methods in
  Natural Language Processing}, pages 2890--2896, Brussels, Belgium,
  October-November 2018. Association for Computational Linguistics.
\newblock \doi{10.18653/v1/D18-1316}.
\newblock URL \url{https://aclanthology.org/D18-1316}.

\bibitem[Athalye et~al.(2018)Athalye, Carlini, and
  Wagner]{athalye2018obfuscated}
Anish Athalye, Nicholas Carlini, and David Wagner.
\newblock Obfuscated gradients give a false sense of security: Circumventing
  defenses to adversarial examples.
\newblock In \emph{International conference on machine learning}, pages
  274--283. PMLR, 2018.

\bibitem[Barreno et~al.(2006)Barreno, Nelson, Sears, Joseph, and
  Tygar]{barreno2006can}
Marco Barreno, Blaine Nelson, Russell Sears, Anthony~D Joseph, and J~Doug
  Tygar.
\newblock Can machine learning be secure?
\newblock In \emph{Proceedings of the 2006 ACM Symposium on Information,
  computer and communications security}, pages 16--25, 2006.

\bibitem[Bradley(1997)]{bradley1997use}
Andrew~P Bradley.
\newblock The use of the area under the roc curve in the evaluation of machine
  learning algorithms.
\newblock \emph{Pattern recognition}, 30\penalty0 (7):\penalty0 1145--1159,
  1997.

\bibitem[Brown et~al.(2020)Brown, Mann, Ryder, Subbiah, Kaplan, Dhariwal,
  Neelakantan, Shyam, Sastry, Askell, Agarwal, Herbert-Voss, Krueger, Henighan,
  Child, Ramesh, Ziegler, Wu, Winter, Hesse, Chen, Sigler, Litwin, Gray, Chess,
  Clark, Berner, McCandlish, Radford, Sutskever, and
  Amodei]{NEURIPS2020_1457c0d6}
Tom Brown, Benjamin Mann, Nick Ryder, Melanie Subbiah, Jared~D Kaplan, Prafulla
  Dhariwal, Arvind Neelakantan, Pranav Shyam, Girish Sastry, Amanda Askell,
  Sandhini Agarwal, Ariel Herbert-Voss, Gretchen Krueger, Tom Henighan, Rewon
  Child, Aditya Ramesh, Daniel Ziegler, Jeffrey Wu, Clemens Winter, Chris
  Hesse, Mark Chen, Eric Sigler, Mateusz Litwin, Scott Gray, Benjamin Chess,
  Jack Clark, Christopher Berner, Sam McCandlish, Alec Radford, Ilya Sutskever,
  and Dario Amodei.
\newblock Language models are few-shot learners.
\newblock In H.~Larochelle, M.~Ranzato, R.~Hadsell, M.F. Balcan, and H.~Lin,
  editors, \emph{Advances in Neural Information Processing Systems}, volume~33,
  pages 1877--1901. Curran Associates, Inc., 2020.
\newblock URL
  \url{https://proceedings.neurips.cc/paper/2020/file/1457c0d6bfcb4967418bfb8ac142f64a-Paper.pdf}.

\bibitem[Carlini and Wagner(2017)]{carlini2017adversarial}
Nicholas Carlini and David Wagner.
\newblock Adversarial examples are not easily detected: Bypassing ten detection
  methods.
\newblock In \emph{Proceedings of the 10th ACM workshop on artificial
  intelligence and security}, pages 3--14, 2017.

\bibitem[Cer et~al.(2018)Cer, Yang, Kong, Hua, Limtiaco, St.~John, Constant,
  Guajardo-Cespedes, Yuan, Tar, Strope, and Kurzweil]{cer-etal-2018-universal}
Daniel Cer, Yinfei Yang, Sheng-yi Kong, Nan Hua, Nicole Limtiaco, Rhomni
  St.~John, Noah Constant, Mario Guajardo-Cespedes, Steve Yuan, Chris Tar,
  Brian Strope, and Ray Kurzweil.
\newblock Universal sentence encoder for {E}nglish.
\newblock In \emph{Proceedings of the 2018 Conference on Empirical Methods in
  Natural Language Processing: System Demonstrations}, pages 169--174,
  Brussels, Belgium, November 2018. Association for Computational Linguistics.
\newblock \doi{10.18653/v1/D18-2029}.
\newblock URL \url{https://aclanthology.org/D18-2029}.

\bibitem[Chapuis et~al.(2020)Chapuis, Colombo, Manica, Labeau, and
  Clavel]{chapuis-etal-2020-hierarchical}
Emile Chapuis, Pierre Colombo, Matteo Manica, Matthieu Labeau, and Chlo{\'e}
  Clavel.
\newblock Hierarchical pre-training for sequence labelling in spoken dialog.
\newblock In \emph{Findings of the Association for Computational Linguistics:
  EMNLP 2020}, pages 2636--2648, Online, November 2020. Association for
  Computational Linguistics.
\newblock \doi{10.18653/v1/2020.findings-emnlp.239}.
\newblock URL \url{https://aclanthology.org/2020.findings-emnlp.239}.

\bibitem[Chen et~al.(2015)Chen, Ting, Washio, and Haffari]{HM}
Bo~Chen, Kai~Ming Ting, Takashi Washio, and Gholamreza Haffari.
\newblock Half-space mass: a maximally robust and efficient data depth method.
\newblock \emph{Machine Learning}, 100\penalty0 (2):\penalty0 677--699, 2015.

\bibitem[Chernozhukov et~al.(2017)Chernozhukov, Galichon, Hallin, and
  Henry]{chernozhukov2017}
Victor Chernozhukov, Alfred Galichon, Marc Hallin, and Marc Henry.
\newblock Monge–kantorovich depth, quantiles, ranks and signs.
\newblock \emph{The Annals of Statistics}, 45\penalty0 (1):\penalty0 223--256,
  02 2017.

\bibitem[Chow(1957)]{chow1957optimum}
Chi-Keung Chow.
\newblock An optimum character recognition system using decision functions.
\newblock \emph{IRE Transactions on Electronic Computers}, \penalty0
  (4):\penalty0 247--254, 1957.

\bibitem[Colombo et~al.(2019)Colombo, Witon, Modi, Kennedy, and
  Kapadia]{colombo-etal-2019-affect}
Pierre Colombo, Wojciech Witon, Ashutosh Modi, James Kennedy, and Mubbasir
  Kapadia.
\newblock Affect-driven dialog generation.
\newblock In \emph{Proceedings of the 2019 Conference of the North {A}merican
  Chapter of the Association for Computational Linguistics: Human Language
  Technologies, Volume 1 (Long and Short Papers)}, pages 3734--3743,
  Minneapolis, Minnesota, June 2019. Association for Computational Linguistics.
\newblock \doi{10.18653/v1/N19-1374}.
\newblock URL \url{https://aclanthology.org/N19-1374}.

\bibitem[Colombo et~al.(2021{\natexlab{a}})Colombo, Chapuis, Labeau, and
  Clavel]{colombo-etal-2021-code}
Pierre Colombo, Emile Chapuis, Matthieu Labeau, and Chlo{\'e} Clavel.
\newblock Code-switched inspired losses for spoken dialog representations.
\newblock In \emph{Proceedings of the 2021 Conference on Empirical Methods in
  Natural Language Processing}, pages 8320--8337, Online and Punta Cana,
  Dominican Republic, November 2021{\natexlab{a}}. Association for
  Computational Linguistics.
\newblock \doi{10.18653/v1/2021.emnlp-main.656}.
\newblock URL \url{https://aclanthology.org/2021.emnlp-main.656}.

\bibitem[Colombo et~al.(2021{\natexlab{b}})Colombo, Chapuis, Labeau, and
  Clavel]{colombo-etal-2021-improving}
Pierre Colombo, Emile Chapuis, Matthieu Labeau, and Chlo{\'e} Clavel.
\newblock Improving multimodal fusion via mutual dependency maximisation.
\newblock In \emph{Proceedings of the 2021 Conference on Empirical Methods in
  Natural Language Processing}, pages 231--245, Online and Punta Cana,
  Dominican Republic, November 2021{\natexlab{b}}. Association for
  Computational Linguistics.
\newblock \doi{10.18653/v1/2021.emnlp-main.21}.
\newblock URL \url{https://aclanthology.org/2021.emnlp-main.21}.

\bibitem[Colombo et~al.(2021{\natexlab{c}})Colombo, Clavel, Yack, and
  Varni]{colombo-etal-2021-beam}
Pierre Colombo, Chlo{\'e} Clavel, Chouchang Yack, and Giovanna Varni.
\newblock Beam search with bidirectional strategies for neural response
  generation.
\newblock In \emph{Proceedings of the 4th International Conference on Natural
  Language and Speech Processing (ICNLSP 2021)}, pages 139--146, Trento, Italy,
  12--13 November 2021{\natexlab{c}}. Association for Computational
  Linguistics.
\newblock URL \url{https://aclanthology.org/2021.icnlsp-1.16}.

\bibitem[Colombo et~al.(2021{\natexlab{d}})Colombo, Piantanida, and
  Clavel]{colombo-etal-2021-novel}
Pierre Colombo, Pablo Piantanida, and Chlo{\'e} Clavel.
\newblock A novel estimator of mutual information for learning to disentangle
  textual representations.
\newblock In \emph{Proceedings of the 59th Annual Meeting of the Association
  for Computational Linguistics and the 11th International Joint Conference on
  Natural Language Processing (Volume 1: Long Papers)}, pages 6539--6550,
  Online, August 2021{\natexlab{d}}. Association for Computational Linguistics.
\newblock \doi{10.18653/v1/2021.acl-long.511}.
\newblock URL \url{https://aclanthology.org/2021.acl-long.511}.

\bibitem[Colombo et~al.(2021{\natexlab{e}})Colombo, Staerman, Clavel, and
  Piantanida]{colombo-etal-2021-automatic}
Pierre Colombo, Guillaume Staerman, Chlo{\'e} Clavel, and Pablo Piantanida.
\newblock Automatic text evaluation through the lens of {W}asserstein
  barycenters.
\newblock In \emph{Proceedings of the 2021 Conference on Empirical Methods in
  Natural Language Processing}, pages 10450--10466, Online and Punta Cana,
  Dominican Republic, November 2021{\natexlab{e}}. Association for
  Computational Linguistics.
\newblock \doi{10.18653/v1/2021.emnlp-main.817}.
\newblock URL \url{https://aclanthology.org/2021.emnlp-main.817}.

\bibitem[Colombo et~al.(2022{\natexlab{a}})Colombo, Dadalto, Staerman, Noiry,
  and Piantanida]{colombo2022beyond}
Pierre Colombo, Eduardo Dadalto, Guillaume Staerman, Nathan Noiry, and Pablo
  Piantanida.
\newblock Beyond mahalanobis distance for textual ood detection.
\newblock \emph{Advances in Neural Information Processing Systems},
  35:\penalty0 17744--17759, 2022{\natexlab{a}}.

\bibitem[Colombo et~al.(2022{\natexlab{b}})Colombo, Noiry, Irurozki, and
  Cl{\'e}men{\c{c}}on]{colombo2022best}
Pierre Colombo, Nathan Noiry, Ekhine Irurozki, and St{\'e}phan
  Cl{\'e}men{\c{c}}on.
\newblock What are the best systems? new perspectives on nlp benchmarking.
\newblock \emph{Advances in Neural Information Processing Systems},
  35:\penalty0 26915--26932, 2022{\natexlab{b}}.

\bibitem[Colombo et~al.(2022{\natexlab{c}})Colombo, Peyrard, Noiry, West, and
  Piantanida]{colombo2022glass}
Pierre Colombo, Maxime Peyrard, Nathan Noiry, Robert West, and Pablo
  Piantanida.
\newblock The glass ceiling of automatic evaluation in natural language
  generation.
\newblock \emph{arXiv preprint arXiv:2208.14585}, 2022{\natexlab{c}}.

\bibitem[Colombo et~al.(2022{\natexlab{d}})Colombo, Staerman, Noiry, and
  Piantanida]{colombo-etal-2022-learning}
Pierre Colombo, Guillaume Staerman, Nathan Noiry, and Pablo Piantanida.
\newblock Learning disentangled textual representations via statistical
  measures of similarity.
\newblock In \emph{Proceedings of the 60th Annual Meeting of the Association
  for Computational Linguistics (Volume 1: Long Papers)}, pages 2614--2630,
  Dublin, Ireland, May 2022{\natexlab{d}}. Association for Computational
  Linguistics.
\newblock \doi{10.18653/v1/2022.acl-long.187}.
\newblock URL \url{https://aclanthology.org/2022.acl-long.187}.

\bibitem[Darrin et~al.(2022)Darrin, Piantanida, and
  Colombo]{darrin2022rainproof}
Maxime Darrin, Pablo Piantanida, and Pierre Colombo.
\newblock Rainproof: An umbrella to shield text generators from
  out-of-distribution data.
\newblock \emph{arXiv preprint arXiv:2212.09171}, 2022.

\bibitem[Darrin et~al.(2023)Darrin, Staerman, Gomes, Cheung, Piantanida, and
  Colombo]{darrin2023unsupervised}
Maxime Darrin, Guillaume Staerman, Eduardo Dadalto~C{\^a}mara Gomes, Jackie~CK
  Cheung, Pablo Piantanida, and Pierre Colombo.
\newblock Unsupervised layer-wise score aggregation for textual ood detection.
\newblock \emph{arXiv preprint arXiv:2302.09852}, 2023.

\bibitem[Davis and Goadrich(2006)]{davis2006relationship}
Jesse Davis and Mark Goadrich.
\newblock The relationship between precision-recall and roc curves.
\newblock In \emph{Proceedings of the 23rd international conference on Machine
  learning}, pages 233--240, 2006.

\bibitem[Devlin et~al.(2019)Devlin, Chang, Lee, and
  Toutanova]{devlin-etal-2019-bert}
Jacob Devlin, Ming-Wei Chang, Kenton Lee, and Kristina Toutanova.
\newblock {BERT}: Pre-training of deep bidirectional transformers for language
  understanding.
\newblock In \emph{Proceedings of the 2019 Conference of the North {A}merican
  Chapter of the Association for Computational Linguistics: Human Language
  Technologies, Volume 1 (Long and Short Papers)}, pages 4171--4186,
  Minneapolis, Minnesota, June 2019. Association for Computational Linguistics.
\newblock \doi{10.18653/v1/N19-1423}.
\newblock URL \url{https://aclanthology.org/N19-1423}.

\bibitem[Dong et~al.(2021)Dong, Luu, Ji, and Liu]{dong2021towards}
Xinshuai Dong, Anh~Tuan Luu, Rongrong Ji, and Hong Liu.
\newblock Towards robustness against natural language word substitutions.
\newblock \emph{arXiv preprint arXiv:2107.13541}, 2021.

\bibitem[Dyckerhoff(2004)]{rainer2004}
Rainer Dyckerhoff.
\newblock Data depth satisfying the projection property.
\newblock \emph{Allgemeines Statistisches Archiv}, 88\penalty0 (2):\penalty0
  163--190, 2004.

\bibitem[Ebrahimi et~al.(2018)Ebrahimi, Rao, Lowd, and
  Dou]{ebrahimi-etal-2018-hotflip}
Javid Ebrahimi, Anyi Rao, Daniel Lowd, and Dejing Dou.
\newblock {H}ot{F}lip: White-box adversarial examples for text classification.
\newblock In \emph{Proceedings of the 56th Annual Meeting of the Association
  for Computational Linguistics (Volume 2: Short Papers)}, pages 31--36,
  Melbourne, Australia, July 2018. Association for Computational Linguistics.
\newblock \doi{10.18653/v1/P18-2006}.
\newblock URL \url{https://aclanthology.org/P18-2006}.

\bibitem[Feinman et~al.(2017)Feinman, Curtin, Shintre, and
  Gardner]{feinman2017detecting}
Reuben Feinman, Ryan~R Curtin, Saurabh Shintre, and Andrew~B Gardner.
\newblock Detecting adversarial samples from artifacts.
\newblock \emph{arXiv preprint arXiv:1703.00410}, 2017.

\bibitem[Feng et~al.(2018)Feng, Wallace, Grissom~II, Iyyer, Rodriguez, and
  Boyd-Graber]{feng-etal-2018-pathologies}
Shi Feng, Eric Wallace, Alvin Grissom~II, Mohit Iyyer, Pedro Rodriguez, and
  Jordan Boyd-Graber.
\newblock Pathologies of neural models make interpretations difficult.
\newblock In \emph{Proceedings of the 2018 Conference on Empirical Methods in
  Natural Language Processing}, pages 3719--3728, Brussels, Belgium,
  October-November 2018. Association for Computational Linguistics.
\newblock \doi{10.18653/v1/D18-1407}.
\newblock URL \url{https://aclanthology.org/D18-1407}.

\bibitem[Gao et~al.(2018)Gao, Lanchantin, Soffa, and Qi]{8424632}
Ji~Gao, Jack Lanchantin, Mary~Lou Soffa, and Yanjun Qi.
\newblock Black-box generation of adversarial text sequences to evade deep
  learning classifiers.
\newblock In \emph{2018 IEEE Security and Privacy Workshops (SPW)}, pages
  50--56, 2018.
\newblock \doi{10.1109/SPW.2018.00016}.

\bibitem[Garg and Ramakrishnan(2020)]{garg-ramakrishnan-2020-bae}
Siddhant Garg and Goutham Ramakrishnan.
\newblock {BAE}: {BERT}-based adversarial examples for text classification.
\newblock In \emph{Proceedings of the 2020 Conference on Empirical Methods in
  Natural Language Processing (EMNLP)}, pages 6174--6181, Online, November
  2020. Association for Computational Linguistics.
\newblock \doi{10.18653/v1/2020.emnlp-main.498}.
\newblock URL \url{https://aclanthology.org/2020.emnlp-main.498}.

\bibitem[Gomes et~al.(2022)Gomes, Alberge, Duhamel, and
  Piantanida]{gomes2022igeood}
Eduardo Dadalto~Camara Gomes, Florence Alberge, Pierre Duhamel, and Pablo
  Piantanida.
\newblock Igeood: An information geometry approach to out-of-distribution
  detection.
\newblock \emph{arXiv preprint arXiv:2203.07798}, 2022.

\bibitem[Goodfellow et~al.(2014)Goodfellow, Shlens, and
  Szegedy]{goodfellow2014explaining}
Ian~J Goodfellow, Jonathon Shlens, and Christian Szegedy.
\newblock Explaining and harnessing adversarial examples.
\newblock \emph{arXiv preprint arXiv:1412.6572}, 2014.

\bibitem[Himmi et~al.(2023)Himmi, Irurozki, Noiry, Clemencon, and
  Colombo]{himmi2023towards}
Anas Himmi, Ekhine Irurozki, Nathan Noiry, Stephan Clemencon, and Pierre
  Colombo.
\newblock Towards more robust nlp system evaluation: Handling missing scores in
  benchmarks.
\newblock \emph{arXiv preprint arXiv:2305.10284}, 2023.

\bibitem[Iyyer et~al.(2018)Iyyer, Wieting, Gimpel, and
  Zettlemoyer]{iyyer-etal-2018-adversarial}
Mohit Iyyer, John Wieting, Kevin Gimpel, and Luke Zettlemoyer.
\newblock Adversarial example generation with syntactically controlled
  paraphrase networks.
\newblock In \emph{Proceedings of the 2018 Conference of the North {A}merican
  Chapter of the Association for Computational Linguistics: Human Language
  Technologies, Volume 1 (Long Papers)}, pages 1875--1885, New Orleans,
  Louisiana, June 2018. Association for Computational Linguistics.
\newblock \doi{10.18653/v1/N18-1170}.
\newblock URL \url{https://aclanthology.org/N18-1170}.

\bibitem[Jia et~al.(2019)Jia, Raghunathan, G{\"o}ksel, and
  Liang]{jia-etal-2019-certified}
Robin Jia, Aditi Raghunathan, Kerem G{\"o}ksel, and Percy Liang.
\newblock Certified robustness to adversarial word substitutions.
\newblock In \emph{Proceedings of the 2019 Conference on Empirical Methods in
  Natural Language Processing and the 9th International Joint Conference on
  Natural Language Processing (EMNLP-IJCNLP)}, pages 4129--4142, Hong Kong,
  China, November 2019. Association for Computational Linguistics.
\newblock \doi{10.18653/v1/D19-1423}.
\newblock URL \url{https://aclanthology.org/D19-1423}.

\bibitem[Jin et~al.(2020)Jin, Jin, Zhou, and Szolovits]{jin2020bert}
Di~Jin, Zhijing Jin, Joey~Tianyi Zhou, and Peter Szolovits.
\newblock Is bert really robust? a strong baseline for natural language attack
  on text classification and entailment.
\newblock In \emph{Proceedings of the AAAI conference on artificial
  intelligence}, volume~34, pages 8018--8025, 2020.

\bibitem[Joachims(1996)]{joachims1996probabilistic}
Thorsten Joachims.
\newblock A probabilistic analysis of the rocchio algorithm with tfidf for text
  categorization.
\newblock Technical report, Carnegie-mellon univ pittsburgh pa dept of computer
  science, 1996.

\bibitem[Johnson(2018)]{johnson2018increasing}
CW~Johnson.
\newblock The increasing risks of risk assessment: On the rise of artificial
  intelligence and non-determinism in safety-critical systems.
\newblock In \emph{the 26th Safety-Critical Systems Symposium}, page~15.
  Safety-Critical Systems Club York, UK., 2018.

\bibitem[Jones et~al.(2020)Jones, Jia, Raghunathan, and Liang]{jones2020robust}
Erik Jones, Robin Jia, Aditi Raghunathan, and Percy Liang.
\newblock Robust encodings: A framework for combating adversarial typos.
\newblock \emph{arXiv preprint arXiv:2005.01229}, 2020.

\bibitem[J{\"o}rnsten(2004)]{jornsten2004clustering}
Rebecka J{\"o}rnsten.
\newblock Clustering and classification based on the l1 data depth.
\newblock \emph{Journal of Multivariate Analysis}, 90\penalty0 (1):\penalty0
  67--89, 2004.

\bibitem[Kamoi and Kobayashi(2020)]{kamoi2020mahalanobis}
Ryo Kamoi and Kei Kobayashi.
\newblock Why is the mahalanobis distance effective for anomaly detection?
\newblock \emph{arXiv preprint arXiv:2003.00402}, 2020.

\bibitem[Kherchouche et~al.(2020)Kherchouche, Fezza, Hamidouche, and
  D{\'e}forges]{kherchouche2020natural}
Anouar Kherchouche, Sid~Ahmed Fezza, Wassim Hamidouche, and Olivier
  D{\'e}forges.
\newblock Natural scene statistics for detecting adversarial examples in deep
  neural networks.
\newblock In \emph{2020 IEEE 22nd International Workshop on Multimedia Signal
  Processing (MMSP)}, pages 1--6. IEEE, 2020.

\bibitem[Kiela et~al.(2021)Kiela, Bartolo, Nie, Kaushik, Geiger, Wu, Vidgen,
  Prasad, Singh, Ringshia, et~al.]{kiela2021dynabench}
Douwe Kiela, Max Bartolo, Yixin Nie, Divyansh Kaushik, Atticus Geiger,
  Zhengxuan Wu, Bertie Vidgen, Grusha Prasad, Amanpreet Singh, Pratik Ringshia,
  et~al.
\newblock Dynabench: Rethinking benchmarking in nlp.
\newblock \emph{arXiv preprint arXiv:2104.14337}, 2021.

\bibitem[Koshevoy and Mosler(1997)]{koshevoy1997}
Gleb Koshevoy and Karl Mosler.
\newblock Zonoid trimming for multivariate distributions.
\newblock \emph{The Annals of Statistics}, 25\penalty0 (5):\penalty0
  1998--2017, 10 1997.

\bibitem[Kuleshov et~al.(2018)Kuleshov, Thakoor, Lau, and
  Ermon]{kuleshov2018adversarial}
Volodymyr Kuleshov, Shantanu Thakoor, Tingfung Lau, and Stefano Ermon.
\newblock Adversarial examples for natural language classification problems.
\newblock 2018.

\bibitem[Lange et~al.(2014)Lange, Mosler, and Mozharovskyi]{LangeMM14}
T.~Lange, K.~Mosler, and P.~Mozharovskyi.
\newblock Dd$\alpha$-classification of asymmetric and fat-tailed data.
\newblock In M.~Spiliopoulou, L.~Schmidt-Thieme, and R.~Janning, editors,
  \emph{Data Analysis, Machine Learning and Knowledge Discovery}, pages 71--78.
  Springer, 2014.

\bibitem[Le et~al.(2021)Le, Park, and Lee]{le-etal-2021-sweet}
Thai Le, Noseong Park, and Dongwon Lee.
\newblock A sweet rabbit hole by {DARCY}: Using honeypots to detect universal
  trigger{'}s adversarial attacks.
\newblock In \emph{Proceedings of the 59th Annual Meeting of the Association
  for Computational Linguistics and the 11th International Joint Conference on
  Natural Language Processing (Volume 1: Long Papers)}, pages 3831--3844,
  Online, August 2021. Association for Computational Linguistics.
\newblock \doi{10.18653/v1/2021.acl-long.296}.
\newblock URL \url{https://aclanthology.org/2021.acl-long.296}.

\bibitem[Lee et~al.(2018)Lee, Lee, Lee, and Shin]{mahalanobis}
Kimin Lee, Kibok Lee, Honglak Lee, and Jinwoo Shin.
\newblock A simple unified framework for detecting out-of-distribution samples
  and adversarial attacks.
\newblock In S.~Bengio, H.~Wallach, H.~Larochelle, K.~Grauman, N.~Cesa-Bianchi,
  and R.~Garnett, editors, \emph{Advances in Neural Information Processing
  Systems 31}, pages 7167--7177. Curran Associates, Inc., 2018.

\bibitem[Levenshtein(1965)]{levenshtein1965leveinshtein}
V~Levenshtein.
\newblock Leveinshtein distance, 1965.

\bibitem[Lhoest et~al.(2021)Lhoest, Villanova~del Moral, Jernite, Thakur, von
  Platen, Patil, Chaumond, Drame, Plu, Tunstall, Davison, {\v{S}}a{\v{s}}ko,
  Chhablani, Malik, Brandeis, Le~Scao, Sanh, Xu, Patry, McMillan-Major, Schmid,
  Gugger, Delangue, Matussi{\`e}re, Debut, Bekman, Cistac, Goehringer, Mustar,
  Lagunas, Rush, and Wolf]{lhoest-etal-2021-datasets}
Quentin Lhoest, Albert Villanova~del Moral, Yacine Jernite, Abhishek Thakur,
  Patrick von Platen, Suraj Patil, Julien Chaumond, Mariama Drame, Julien Plu,
  Lewis Tunstall, Joe Davison, Mario {\v{S}}a{\v{s}}ko, Gunjan Chhablani,
  Bhavitvya Malik, Simon Brandeis, Teven Le~Scao, Victor Sanh, Canwen Xu,
  Nicolas Patry, Angelina McMillan-Major, Philipp Schmid, Sylvain Gugger,
  Cl{\'e}ment Delangue, Th{\'e}o Matussi{\`e}re, Lysandre Debut, Stas Bekman,
  Pierric Cistac, Thibault Goehringer, Victor Mustar, Fran{\c{c}}ois Lagunas,
  Alexander Rush, and Thomas Wolf.
\newblock Datasets: A community library for natural language processing.
\newblock In \emph{Proceedings of the 2021 Conference on Empirical Methods in
  Natural Language Processing: System Demonstrations}, pages 175--184, Online
  and Punta Cana, Dominican Republic, November 2021. Association for
  Computational Linguistics.
\newblock URL \url{https://aclanthology.org/2021.emnlp-demo.21}.

\bibitem[Li et~al.(2021)Li, Zhang, Peng, Chen, Brockett, Sun, and
  Dolan]{li-etal-2021-contextualized}
Dianqi Li, Yizhe Zhang, Hao Peng, Liqun Chen, Chris Brockett, Ming-Ting Sun,
  and Bill Dolan.
\newblock Contextualized perturbation for textual adversarial attack.
\newblock In \emph{Proceedings of the 2021 Conference of the North American
  Chapter of the Association for Computational Linguistics: Human Language
  Technologies}, pages 5053--5069, Online, June 2021. Association for
  Computational Linguistics.
\newblock \doi{10.18653/v1/2021.naacl-main.400}.
\newblock URL \url{https://aclanthology.org/2021.naacl-main.400}.

\bibitem[Li et~al.(2018)Li, Ji, Du, Li, and Wang]{li2018textbugger}
Jinfeng Li, Shouling Ji, Tianyu Du, Bo~Li, and Ting Wang.
\newblock Textbugger: Generating adversarial text against real-world
  applications.
\newblock \emph{arXiv preprint arXiv:1812.05271}, 2018.

\bibitem[Li et~al.(2020)Li, Ma, Guo, Xue, and Qiu]{li-etal-2020-bert-attack}
Linyang Li, Ruotian Ma, Qipeng Guo, Xiangyang Xue, and Xipeng Qiu.
\newblock {BERT}-{ATTACK}: Adversarial attack against {BERT} using {BERT}.
\newblock In \emph{Proceedings of the 2020 Conference on Empirical Methods in
  Natural Language Processing (EMNLP)}, pages 6193--6202, Online, November
  2020. Association for Computational Linguistics.
\newblock \doi{10.18653/v1/2020.emnlp-main.500}.
\newblock URL \url{https://aclanthology.org/2020.emnlp-main.500}.

\bibitem[Liu(1992)]{Liu92}
Regina~Y. Liu.
\newblock \emph{Data Depth and Multivariate Rank Tests}, page 279–294.
\newblock North-Holland, Amsterdam, 1992.

\bibitem[Liu et~al.(2019)Liu, Ott, Goyal, Du, Joshi, Chen, Levy, Lewis,
  Zettlemoyer, and Stoyanov]{liu2019roberta}
Yinhan Liu, Myle Ott, Naman Goyal, Jingfei Du, Mandar Joshi, Danqi Chen, Omer
  Levy, Mike Lewis, Luke Zettlemoyer, and Veselin Stoyanov.
\newblock Roberta: A robustly optimized bert pretraining approach.
\newblock \emph{arXiv preprint arXiv:1907.11692}, 2019.

\bibitem[Ma et~al.(2018)Ma, Li, Wang, Erfani, Wijewickrema, Schoenebeck, Song,
  Houle, and Bailey]{ma2018characterizing}
Xingjun Ma, Bo~Li, Yisen Wang, Sarah~M Erfani, Sudanthi Wijewickrema, Grant
  Schoenebeck, Dawn Song, Michael~E Houle, and James Bailey.
\newblock Characterizing adversarial subspaces using local intrinsic
  dimensionality.
\newblock \emph{arXiv preprint arXiv:1801.02613}, 2018.

\bibitem[Maas et~al.(2011)Maas, Daly, Pham, Huang, Ng, and
  Potts]{maas-etal-2011-learning}
Andrew~L. Maas, Raymond~E. Daly, Peter~T. Pham, Dan Huang, Andrew~Y. Ng, and
  Christopher Potts.
\newblock Learning word vectors for sentiment analysis.
\newblock In \emph{Proceedings of the 49th Annual Meeting of the Association
  for Computational Linguistics: Human Language Technologies}, pages 142--150,
  Portland, Oregon, USA, June 2011. Association for Computational Linguistics.
\newblock URL \url{https://aclanthology.org/P11-1015}.

\bibitem[Mahalanobis(1936)]{mahalanobis1936generalized}
Prasanta~Chandra Mahalanobis.
\newblock On the generalized distance in statistics.
\newblock National Institute of Science of India, 1936.

\bibitem[Mikolov et~al.(2013)Mikolov, Chen, Corrado, and
  Dean]{mikolov2013efficient}
Tomas Mikolov, Kai Chen, Greg Corrado, and Jeffrey Dean.
\newblock Efficient estimation of word representations in vector space.
\newblock \emph{arXiv preprint arXiv:1301.3781}, 2013.

\bibitem[Miller(1995)]{miller1995wordnet}
George~A Miller.
\newblock Wordnet: a lexical database for english.
\newblock \emph{Communications of the ACM}, 38\penalty0 (11):\penalty0 39--41,
  1995.

\bibitem[Miller et~al.(1990)Miller, Beckwith, Fellbaum, Gross, and
  Miller]{miller1990introduction}
George~A Miller, Richard Beckwith, Christiane Fellbaum, Derek Gross, and
  Katherine~J Miller.
\newblock Introduction to wordnet: An on-line lexical database.
\newblock \emph{International journal of lexicography}, 3\penalty0
  (4):\penalty0 235--244, 1990.

\bibitem[Morris et~al.(2020{\natexlab{a}})Morris, Lifland, Lanchantin, Ji, and
  Qi]{morris-etal-2020-reevaluating}
John Morris, Eli Lifland, Jack Lanchantin, Yangfeng Ji, and Yanjun Qi.
\newblock Reevaluating adversarial examples in natural language.
\newblock In \emph{Findings of the Association for Computational Linguistics:
  EMNLP 2020}, pages 3829--3839, Online, November 2020{\natexlab{a}}.
  Association for Computational Linguistics.
\newblock \doi{10.18653/v1/2020.findings-emnlp.341}.
\newblock URL \url{https://aclanthology.org/2020.findings-emnlp.341}.

\bibitem[Morris et~al.(2020{\natexlab{b}})Morris, Lifland, Yoo, Grigsby, Jin,
  and Qi]{morris2020textattack}
John Morris, Eli Lifland, Jin~Yong Yoo, Jake Grigsby, Di~Jin, and Yanjun Qi.
\newblock Textattack: A framework for adversarial attacks, data augmentation,
  and adversarial training in nlp.
\newblock In \emph{Proceedings of the 2020 Conference on Empirical Methods in
  Natural Language Processing: System Demonstrations}, pages 119--126,
  2020{\natexlab{b}}.

\bibitem[Mosler(2013)]{Mosler13}
Karl. Mosler.
\newblock Depth statistics.
\newblock In C.~Becker, R.~Fried, and S.~Kuhnt, editors, \emph{Robustness and
  Complex Data Structures: Festschrift in Honour of Ursula Gather}, pages
  17--34. Springer, 2013.

\bibitem[Mosler and Mozharovskyi(2020)]{mosler2020choosing}
Karl Mosler and Pavlo Mozharovskyi.
\newblock Choosing among notions of multivariate depth statistics.
\newblock \emph{arXiv preprint arXiv:2004.01927}, 2020.

\bibitem[Mozes et~al.(2021)Mozes, Stenetorp, Kleinberg, and
  Griffin]{mozes-etal-2021-frequency}
Maximilian Mozes, Pontus Stenetorp, Bennett Kleinberg, and Lewis Griffin.
\newblock Frequency-guided word substitutions for detecting textual adversarial
  examples.
\newblock In \emph{Proceedings of the 16th Conference of the European Chapter
  of the Association for Computational Linguistics: Main Volume}, pages
  171--186, Online, April 2021. Association for Computational Linguistics.
\newblock URL \url{https://www.aclweb.org/anthology/2021.eacl-main.13}.

\bibitem[Peyré and Cuturi(2019)]{Peyre}
Gabriel Peyré and Marco Cuturi.
\newblock Computational optimal transport.
\newblock \emph{Foundations and Trends® in Machine Learning}, 11\penalty0
  (5-6):\penalty0 355--607, 2019.

\bibitem[Pichler et~al.(2022)Pichler, Colombo, Boudiaf, Koliander, and
  Piantanida]{pichler2022differential}
Georg Pichler, Pierre Jean~A Colombo, Malik Boudiaf, G{\"u}nther Koliander, and
  Pablo Piantanida.
\newblock A differential entropy estimator for training neural networks.
\newblock In \emph{International Conference on Machine Learning}, pages
  17691--17715. PMLR, 2022.

\bibitem[Picot et~al.(2022{\natexlab{a}})Picot, Noiry, Piantanida, and
  Colombo]{picot2022adversarial}
Marine Picot, Nathan Noiry, Pablo Piantanida, and Pierre Colombo.
\newblock Adversarial attack detection under realistic constraints.
\newblock 2022{\natexlab{a}}.

\bibitem[Picot et~al.(2022{\natexlab{b}})Picot, Staerman, Granese, Noiry,
  Messina, Piantanida, and Colombo]{picot2022simple}
Marine Picot, Guillaume Staerman, Federica Granese, Nathan Noiry, Francisco
  Messina, Pablo Piantanida, and Pierre Colombo.
\newblock A simple unsupervised data depth-based method to detect adversarial
  images.
\newblock 2022{\natexlab{b}}.

\bibitem[Podolskiy et~al.(2021)Podolskiy, Lipin, Bout, Artemova, and
  Piontkovskaya]{podolskiy2021revisiting}
Alexander Podolskiy, Dmitry Lipin, Andrey Bout, Ekaterina Artemova, and Irina
  Piontkovskaya.
\newblock Revisiting mahalanobis distance for transformer-based out-of-domain
  detection.
\newblock \emph{arXiv preprint arXiv:2101.03778}, 2021.

\bibitem[Pruthi et~al.(2019)Pruthi, Dhingra, and
  Lipton]{pruthi-etal-2019-combating}
Danish Pruthi, Bhuwan Dhingra, and Zachary~C. Lipton.
\newblock Combating adversarial misspellings with robust word recognition.
\newblock In \emph{Proceedings of the 57th Annual Meeting of the Association
  for Computational Linguistics}, pages 5582--5591, Florence, Italy, July 2019.
  Association for Computational Linguistics.
\newblock \doi{10.18653/v1/P19-1561}.
\newblock URL \url{https://aclanthology.org/P19-1561}.

\bibitem[Ramsay et~al.(2019)Ramsay, Durocher, and Leblanc]{IRW}
Kelly Ramsay, Stéphane Durocher, and Alexandre Leblanc.
\newblock Integrated rank-weighted depth.
\newblock \emph{Journal of Multivariate Analysis}, 173:\penalty0 51--69, 2019.

\bibitem[Ren et~al.(2021)Ren, Fort, Liu, Roy, Padhy, and
  Lakshminarayanan]{ren2021simple}
Jie Ren, Stanislav Fort, Jeremiah Liu, Abhijit~Guha Roy, Shreyas Padhy, and
  Balaji Lakshminarayanan.
\newblock A simple fix to mahalanobis distance for improving near-ood
  detection.
\newblock \emph{arXiv preprint arXiv:2106.09022}, 2021.

\bibitem[Ren et~al.(2019)Ren, Deng, He, and Che]{ren-etal-2019-generating}
Shuhuai Ren, Yihe Deng, Kun He, and Wanxiang Che.
\newblock Generating natural language adversarial examples through probability
  weighted word saliency.
\newblock In \emph{Proceedings of the 57th Annual Meeting of the Association
  for Computational Linguistics}, pages 1085--1097, Florence, Italy, July 2019.
  Association for Computational Linguistics.
\newblock \doi{10.18653/v1/P19-1103}.
\newblock URL \url{https://aclanthology.org/P19-1103}.

\bibitem[Ribeiro et~al.(2020)Ribeiro, Wu, Guestrin, and
  Singh]{ribeiro-etal-2020-beyond}
Marco~Tulio Ribeiro, Tongshuang Wu, Carlos Guestrin, and Sameer Singh.
\newblock Beyond accuracy: Behavioral testing of {NLP} models with
  {C}heck{L}ist.
\newblock In \emph{Proceedings of the 58th Annual Meeting of the Association
  for Computational Linguistics}, pages 4902--4912, Online, July 2020.
  Association for Computational Linguistics.
\newblock \doi{10.18653/v1/2020.acl-main.442}.
\newblock URL \url{https://aclanthology.org/2020.acl-main.442}.

\bibitem[Sastry and Oore(2020)]{sastry2020detecting}
Chandramouli~Shama Sastry and Sageev Oore.
\newblock Detecting out-of-distribution examples with gram matrices.
\newblock In \emph{International Conference on Machine Learning}, pages
  8491--8501. PMLR, 2020.

\bibitem[Socher et~al.(2013)Socher, Perelygin, Wu, Chuang, Manning, Ng, and
  Potts]{socher-etal-2013-recursive}
Richard Socher, Alex Perelygin, Jean Wu, Jason Chuang, Christopher~D. Manning,
  Andrew Ng, and Christopher Potts.
\newblock Recursive deep models for semantic compositionality over a sentiment
  treebank.
\newblock In \emph{Proceedings of the 2013 Conference on Empirical Methods in
  Natural Language Processing}, pages 1631--1642, Seattle, Washington, USA,
  October 2013. Association for Computational Linguistics.
\newblock URL \url{https://aclanthology.org/D13-1170}.

\bibitem[Staerman(2022)]{phdguigui}
Guillaume Staerman.
\newblock \emph{Functional anomaly detection and robust estimation}.
\newblock PhD thesis, Institut polytechnique de Paris, 2022.

\bibitem[Staerman et~al.(2020)Staerman, Mozharovskyi, and
  Cl\'emençon]{staerman2020}
Guillaume Staerman, Pavlo Mozharovskyi, and St\'ephan Cl\'emençon.
\newblock The area of the convex hull of sampled curves: a robust functional
  statistical depth measure.
\newblock In \emph{Proceedings of the 23nd International Conference on
  Artificial Intelligence and Statistics}, volume 108, pages 570--579, 2020.

\bibitem[Staerman et~al.(2021{\natexlab{a}})Staerman, Mozharovskyi, and
  Clémençon]{AIIRW}
Guillaume Staerman, Pavlo Mozharovskyi, and Stéphan Clémençon.
\newblock Affine-invariant integrated rank-weighted depth: Definition,
  properties and finite sample analysis.
\newblock \emph{arXiv preprint arXiv:2106.11068}, 2021{\natexlab{a}}.

\bibitem[Staerman et~al.(2021{\natexlab{b}})Staerman, Mozharovskyi, Colombo,
  Cl{\'e}men{\c c}on, and d'Alch{\'e} Buc]{dr_distance}
Guillaume Staerman, Pavlo Mozharovskyi, Pierre Colombo, St{\'e}phan
  Cl{\'e}men{\c c}on, and Florence d'Alch{\'e} Buc.
\newblock A pseudo-metric between probability distributions based on
  depth-trimmed regions.
\newblock \emph{arXiv preprint arXiv:2103.12711}, 2021{\natexlab{b}}.

\bibitem[Staerman et~al.(2022)Staerman, Adjakossa, Mozharovskyi, Hofer, Gupta,
  and Cl{\'e}men{\c{c}}on]{benchmarkfad}
Guillaume Staerman, Eric Adjakossa, Pavlo Mozharovskyi, Vera Hofer, Jayant~Sen
  Gupta, and Stephan Cl{\'e}men{\c{c}}on.
\newblock Functional anomaly detection: a benchmark study.
\newblock \emph{arXiv preprint arXiv:2201.05115}, 2022.

\bibitem[Subbaswamy and Saria(2020)]{adarsh_health_ai_2020}
Adarsh Subbaswamy and Suchi Saria.
\newblock From development to deployment: dataset shift, causality, and
  shift-stable models in health ai.
\newblock \emph{Biostatistics}, 21\penalty0 (2):\penalty0 345--352, April 2020.
\newblock ISSN 1465-4644.
\newblock \doi{10.1093/biostatistics/kxz041}.

\bibitem[Szegedy et~al.(2013)Szegedy, Zaremba, Sutskever, Bruna, Erhan,
  Goodfellow, and Fergus]{szegedy2013intriguing}
Christian Szegedy, Wojciech Zaremba, Ilya Sutskever, Joan Bruna, Dumitru Erhan,
  Ian Goodfellow, and Rob Fergus.
\newblock Intriguing properties of neural networks.
\newblock \emph{arXiv preprint arXiv:1312.6199}, 2013.

\bibitem[Szegedy et~al.(2014)Szegedy, Zaremba, Sutskever, Bruna, Erhan,
  Goodfellow, and Fergus]{Szegedy2014ICLR}
Christian Szegedy, Wojciech Zaremba, Ilya Sutskever, Joan Bruna, Dumitru Erhan,
  Ian~J. Goodfellow, and Rob Fergus.
\newblock Intriguing properties of neural networks.
\newblock In \emph{2nd International Conference on Learning Representations},
  2014.

\bibitem[Tramer et~al.(2020)Tramer, Carlini, Brendel, and
  Madry]{tramer2020adaptive}
Florian Tramer, Nicholas Carlini, Wieland Brendel, and Aleksander Madry.
\newblock On adaptive attacks to adversarial example defenses.
\newblock \emph{Advances in Neural Information Processing Systems},
  33:\penalty0 1633--1645, 2020.

\bibitem[Tukey(1975)]{Tukey75}
John~W. Tukey.
\newblock Mathematics and the picturing of data.
\newblock In \emph{Proceedings of the International Congress of
  Mathematicians}, volume~2, pages 523--531, 1975.

\bibitem[Wang et~al.(2020)Wang, Wang, Cheng, Gan, Jia, Li, and
  Liu]{wang2020infobert}
Boxin Wang, Shuohang Wang, Yu~Cheng, Zhe Gan, Ruoxi Jia, Bo~Li, and Jingjing
  Liu.
\newblock Infobert: Improving robustness of language models from an information
  theoretic perspective.
\newblock \emph{arXiv preprint arXiv:2010.02329}, 2020.

\bibitem[Wang et~al.(2019)Wang, Jin, and He]{wang2019natural}
Xiaosen Wang, Hao Jin, and Kun He.
\newblock Natural language adversarial attack and defense in word level.
\newblock 2019.

\bibitem[Wolf et~al.(2020)Wolf, Debut, Sanh, Chaumond, Delangue, Moi, Cistac,
  Rault, Louf, Funtowicz, Davison, Shleifer, von Platen, Ma, Jernite, Plu, Xu,
  Le~Scao, Gugger, Drame, Lhoest, and Rush]{wolf-etal-2020-transformers}
Thomas Wolf, Lysandre Debut, Victor Sanh, Julien Chaumond, Clement Delangue,
  Anthony Moi, Pierric Cistac, Tim Rault, Remi Louf, Morgan Funtowicz, Joe
  Davison, Sam Shleifer, Patrick von Platen, Clara Ma, Yacine Jernite, Julien
  Plu, Canwen Xu, Teven Le~Scao, Sylvain Gugger, Mariama Drame, Quentin Lhoest,
  and Alexander Rush.
\newblock Transformers: State-of-the-art natural language processing.
\newblock In \emph{Proceedings of the 2020 Conference on Empirical Methods in
  Natural Language Processing: System Demonstrations}, pages 38--45, Online,
  October 2020. Association for Computational Linguistics.
\newblock \doi{10.18653/v1/2020.emnlp-demos.6}.
\newblock URL \url{https://aclanthology.org/2020.emnlp-demos.6}.

\bibitem[Yoo and Qi(2021)]{yoo-qi-2021-towards-improving}
Jin~Yong Yoo and Yanjun Qi.
\newblock Towards improving adversarial training of {NLP} models.
\newblock In \emph{Findings of the Association for Computational Linguistics:
  EMNLP 2021}, pages 945--956, Punta Cana, Dominican Republic, November 2021.
  Association for Computational Linguistics.
\newblock \doi{10.18653/v1/2021.findings-emnlp.81}.
\newblock URL \url{https://aclanthology.org/2021.findings-emnlp.81}.

\bibitem[Yoo et~al.(2022)Yoo, Kim, Jang, and Kwak]{yoo2022detection}
KiYoon Yoo, Jangho Kim, Jiho Jang, and Nojun Kwak.
\newblock Detection of word adversarial examples in text classification:
  Benchmark and baseline via robust density estimation.
\newblock \emph{arXiv preprint arXiv:2203.01677}, 2022.

\bibitem[Zang et~al.(2019)Zang, Qi, Yang, Liu, Zhang, Liu, and
  Sun]{zang2019word}
Yuan Zang, Fanchao Qi, Chenghao Yang, Zhiyuan Liu, Meng Zhang, Qun Liu, and
  Maosong Sun.
\newblock Word-level textual adversarial attacking as combinatorial
  optimization.
\newblock \emph{arXiv preprint arXiv:1910.12196}, 2019.

\bibitem[Zeng et~al.(2021)Zeng, Qi, Zhou, Zhang, Ma, Hou, Zang, Liu, and
  Sun]{zeng-etal-2021-openattack}
Guoyang Zeng, Fanchao Qi, Qianrui Zhou, Tingji Zhang, Zixian Ma, Bairu Hou,
  Yuan Zang, Zhiyuan Liu, and Maosong Sun.
\newblock {O}pen{A}ttack: An open-source textual adversarial attack toolkit.
\newblock In \emph{Proceedings of the 59th Annual Meeting of the Association
  for Computational Linguistics and the 11th International Joint Conference on
  Natural Language Processing: System Demonstrations}, pages 363--371, Online,
  August 2021. Association for Computational Linguistics.
\newblock \doi{10.18653/v1/2021.acl-demo.43}.
\newblock URL \url{https://aclanthology.org/2021.acl-demo.43}.

\bibitem[Zhang* et~al.(2020)Zhang*, Kishore*, Wu*, Weinberger, and
  Artzi]{bert-score}
Tianyi Zhang*, Varsha Kishore*, Felix Wu*, Kilian~Q. Weinberger, and Yoav
  Artzi.
\newblock Bertscore: Evaluating text generation with bert.
\newblock In \emph{International Conference on Learning Representations}, 2020.
\newblock URL \url{https://openreview.net/forum?id=SkeHuCVFDr}.

\bibitem[Zhou et~al.(2021)Zhou, Zheng, Hsieh, Chang, and
  Huang]{zhou2021defense}
Yi~Zhou, Xiaoqing Zheng, Cho-Jui Hsieh, Kai-Wei Chang, and Xuan-Jing Huang.
\newblock Defense against synonym substitution-based adversarial attacks via
  dirichlet neighborhood ensemble.
\newblock In \emph{Proceedings of the 59th Annual Meeting of the Association
  for Computational Linguistics and the 11th International Joint Conference on
  Natural Language Processing (Volume 1: Long Papers)}, pages 5482--5492, 2021.

\bibitem[Zuo and Serfling(2000)]{ZuoSerfling00}
Yijun Zuo and Robert Serfling.
\newblock General notions of statistical depth function.
\newblock \emph{The Annals of Statistics}, 28\penalty0 (2):\penalty0 461--482,
  2000.

\end{thebibliography}
\clearpage
\appendixpage
\startcontents[sections]
\printcontents[sections]{l}{1}{\setcounter{tocdepth}{2}}

\section{Approximation algorithms}\label{sec:algo}

In this section, we present the algorithms, originally proposed in \cite{HM} and adapted to our problem, to approximate $D_{HM}$ that is used in {\DETECTOR} (see Algorithm \ref{algo:train} for the training and Algorithm \ref{algo:test} for the testing).  

\subsection{Computational aspects}\label{appendix:harcore}
The first step is to draw uniformly at random a set of closed halfspaces of $\mathbb{R}^d$. Drawing a halfspace is equivalent to drawing a hyperplane, which is accomplished by sampling a direction $u$ from the unit hypersphere $\mathbb{S}^{d-1}$ as well as a threshold ruled by a hyperparameter $\lambda$ that both uniquely define a hyperplane/halfspace. To avoid halfspaces carrying no information, thresholds are chosen such that at least one training sample belongs to each of the constructed halfspaces requiring to project the training set on $u$. This procedure is repeated $K\geq 1$ times, where $K$ is chosen by the user as is $\lambda$. The computation can be accelerated by performing this procedure on a subsample of the training set. The training part, which is outlined in Algorithm~\ref{algo:train} in Appendix~\ref{sec:algo}, allows to obtain $K$ closed halfspaces as well as their $K$ complementary spaces leading to   $2K$ halfspaces.  The test part lie in evaluating the depth score of any new observation $\mathbf{x} \in \mathbb{R}^d$ by using the pre-defined halfspaces. Indeed, $\mathbf{x}$ belongs to $K$ among the $2K$ constructed halfspaces during the training step.  The goal is then to compute the probability mass, \textit{i.e.}, the proportion of training samples in the $K$ halfspaces to which $\mathbf{x}$ belongs and then, compute the mean values of these proportions. This testing step is outlined in the Algorithm~\ref{algo:test} in Appendix~\ref{sec:algo}.

\begin{algorithm}[!htb]
\label{algo:train}
\caption{Training algorithm for the approximation of $D_{\mathrm{HM}}$.}\label{algo:train}
\INPUT{:}  sample $\mathcal{D}_{y}=\{f_{\psi}^{L}(\mathbf{x}_i) \; : \;  y_i=y \}$.

\vspace{0.1cm}

\INIT{:} Number of halfspaces $K$; sub-sample size $n_s$; hyperparameter $\lambda$.
      \begin{algorithmic}[1]
      \For{ $k=1,\ldots, K$}

      \vspace{0.2cm}
      
      \State Draw  $\mathcal{D}_{y,n_s}$, a sub-sample of $\mathcal{D}_{y}$ with size $n_s$ without replacement.
      
      \vspace{0.2cm}
      \State Draw randomly and uniformly a direction $u_k$ in $\mathbb{S}^{d-1}$.

      \vspace{0.2cm}
      
      \State Compute   $\langle u_k, f_{\psi}^{L}(\mathbf{x}_i) \rangle$ for every $f_{\psi}^{L}(\mathbf{x}_i) \in \mathcal{D}_{y,n_s}$  such that  $p_{k,i} \triangleq\langle u_k, f_{\psi}^{L}(\mathbf{x}_i)  \rangle$.
      \vspace{0.2cm}
      
      \State Set $\mathrm{mid}_k= \big (\underset{i}{\max} ~p_{k,i} + \underset{i}{\min} ~p_{k,i} \big ) / 2$ and $\mathrm{range}_k= \underset{i}{\max} ~p_{k,i} - \underset{i}{\min} ~p_{k,i}$.

      \vspace{0.2cm}
      
      \State Randomly and uniformly select $\kappa_k$ in $\big [\mathrm{mid}_k - \frac{\lambda}{2} \mathrm{range}_k, \;\mathrm{mid}_k + \frac{\lambda}{2} \mathrm{range}_k \big ]$.

      \vspace{0.35cm} 
      
      \State Set  $m_{k}^{\mathrm{left}} \hspace{-0.1cm}=  \dfrac{| \{\mathbf{z} \in \mathcal{D}_{y,n_s}: \; p_{k,i} < \kappa_k \}  |} {n_s}$ and $m_{k}^{\mathrm{right}} \hspace{-0.1cm}= \dfrac{| \{\mathbf{z} \in \mathcal{D}_{y,n_s}: \; p_{k,i} \geq \kappa_k \}  |} {n_s}$.

      \vspace{0.2cm}

\EndFor 
      \end{algorithmic}
      \OUTPUT{:} $\{u_k, \kappa_k, m_{k}^{\mathrm{left}}, m_{k}^{\mathrm{right}} \}_{k=1}^{K}$.
\end{algorithm}

\begin{algorithm}[!htb]
\label{algo:test}
\caption{Testing algorithm for the approximation of $D_{\mathrm{HM}}$.}\label{algo:test}
\INPUT{:}  test observation $f_{\psi}^{L}(\mathbf{x})$; $\{u_k, \kappa_k, m_{k}^{\mathrm{left}}, m_{k}^{\mathrm{right}} \}_{k=1}^{K}$.

\vspace{0.1cm}

\INIT{:} HM=0.
      \begin{algorithmic}[1]

      \For{ $k=1,\ldots, K$} 
      
      \vspace{0.2cm}
      \State Project $f_{\psi}^{L}(\mathbf{x})$ onto $u_k$ and such that $p_k^\ell = \langle f_{\psi}^{L}(\mathbf{x}),u_k \rangle$.

      \vspace{0.2cm}
      
      \State HM = HM + $m_{k}^{\mathrm{left}}~\mathbb{I}(p_k^\ell <\kappa_k)$ + $m_{k}^{\mathrm{right}}~\mathbb{I}(p_k^\ell\geq \kappa_k)$.

      \vspace{0.2cm}
\EndFor 
      \end{algorithmic}
      \OUTPUT{:} $D_{\mathrm{HM}}(f_{\psi}^{L}(\mathbf{x}),\mathcal{D}_{y})=\text{HM} /K$.
\end{algorithm}

\section{Additional Details on {\BENCHMARK} Construction}

For completeness, we regroup in \autoref{tab:benchmark_considered_attacks} additional details on the attacks used in {\BENCHMARK}.

In this paper we focused our evaluation on existing attacks. In the future, a possible extension of {\BENCHMARK} would be to use the methodology of Dynabench \cite{kiela2021dynabench} and rely on human feedback to attack the model.

\begin{table*}[!ht]
 \caption{Considered attacks for {\BENCHMARK} construction.}
 \label{tab:benchmark_considered_attacks}
 \centering
 \resizebox{\textwidth}{!}{\begin{tabular}{ccp{4cm}p{4.5cm}}\hline
 Full Name & Acronym &Idea & Type of Constraint \\\hline
pruthi \cite{pruthi-etal-2019-combating}& PRU & Simulation of common typos using greedy search for untargeted classification & Minimum word length, maximum number of words perturbed\\
 textbugger \cite{li2018textbugger}& TB& Character-based attack (\textit{i.e} swap, deletion, subtitution) &Cosine with USE \cite{cer-etal-2018-universal} \\
 iga \cite{wang2019natural} &IG & Genetic algorithm to perform word substitution& Percentage of perturbed words  and word embedding distance on Word2Vect \cite{mikolov2013efficient}\\
 deepwordbug \cite{8424632} &DWB &Character-based attack (\textit{i.e} swap, deletion, subtitution) & Levenshtein distance \cite{levenshtein1965leveinshtein}\\
 kuleshov \cite{kuleshov2018adversarial} &KUL &Attack using embedding swap & Cosine and language model similarity \\
 clare\cite{li-etal-2021-contextualized} &CLA & Attack using token insertion, merge and swap & Embedding similarity\\
bae \cite{garg-ramakrishnan-2020-bae} &BAE& Attack using BERT MLM combined with a greedy search & Number of perturbed words and cosine with USE \cite{cer-etal-2018-universal} \\
pwws \cite{ren-etal-2019-generating}&PWWS & Word swap based on WordNet synonyms &\\
 textfooler \cite{jin2020bert}&TF & Attack using embedding swap& Embedding similarity and POS match with word and embedding swap \\
TF-adjusted\cite{morris-etal-2020-reevaluating} &TF-ADF& Attack using embedding swap & USE and word embedding similarity \\
input-reduction \cite{feng-etal-2018-pathologies} &IR& Greedy attack using word importance ranking via greedy search &\\ 
 checklist \cite{ribeiro-etal-2020-beyond}&CHK &Using contraction/extension and changing numbers, locations, names&\\
 \hline
 \end{tabular}}
\end{table*}

\section{Additional Results}
This section gathers additional experimental results to allow the curious reader to draw fine conclusions. Formally, we conduct:
\begin{itemize}
    \item a detailed analysis of the detectors' performances per attack (see \autoref{ssec:analyse_per_attack_sum}).
    \item a detailed analysis of the detectors' performances per dataset across all the considered metrics (see \autoref{ssec:analyse_per_ds_sum}).
    \item the extended figures of \autoref{ssec:identifying_key} (see~\autoref{ssec:analyse_per_pretrained_sum}).
    \item an analysis per dataset/per attack of the different detector performances on {\BENCHMARK} in \autoref{ssec:analyse_per_pretrained_per_attack_sum}.
    \item an comparative study of the detector's performance between semantic and syntactic attacks.
    \item an in-depth reflection on the possibility of building multi-layer detectors (see~\autoref{ssec:future_rd}).
\end{itemize}

\subsection{Fine grained analysis per attack}\label{ssec:analyse_per_attack_sum}
In \autoref{tab:average_per_model},  we report the average performances on {\BENCHMARK} for each detector on each model under each attack's threat. First, it is interesting to note that {\DETECTOR} strongly outperforms other methods on most of the configurations. Then, corroborating previous observations, we find that changing attacks, encoders, and metrics largely influence the detection performances.  
\\\noindent \textbf{Takeaways.} These findings validate our extended {\BENCHMARK}, as in real-life scenario, practitioners need to ensure that the detection methods works well on a large number of attacks for different types of models.
\begin{table*}[!ht]
    \centering
        \caption{Average performances on {\BENCHMARK} per model and per attack}
    \label{tab:average_per_model}
\resizebox{!}{0.5\textwidth}{\begin{tabular}{cccccccc}  \toprule     
&                    &        &  {\AUROC} &  {\FPR} &{\AUPRIN} &{\AUPROUT}&{\ERR} \\
\midrule      
\texttt{BERT} & $D_{M}$ & BAE   &   \result{87.5} {        7.6 } & \result{ 39.3 }{     20.3 }&     \result{88.5} {          8.8 }&     \result{ 84.1 }{           8.0 }& \result{  24.6 }{       10.1 }\\ 
&                    & DWB      &   \result{88.6} {        6.8 } & \result{ 60.0 }{     24.3 }&     \result{90.8} {          5.7 }&     \result{ 84.7 }{           9.2 }& \result{  32.4 }{       12.1 }\\ 
&                    & IG       &   \result{87.8} {        5.0 } & \result{ 72.0 }{     15.2 }&     \result{90.7} {          4.1 }&     \result{ 82.3 }{           7.2 }& \result{  38.2 }{        7.7 }\\          
&                    & KUL      &   \result{88.2} {        4.7 } & \result{ 67.4 }{     17.6 }&     \result{90.9} {          4.0 }&     \result{ 83.1 }{           6.9 }& \result{  35.8 }{        8.9 }\\  
&                    & PRU      &   \result{84.6} {        8.6 } & \result{ 67.0 }{     21.8 }&     \result{86.9} {          7.8 }&     \result{ 80.5 }{          10.2 }& \result{  35.7 }{       10.7 }\\ 
&                    & PWWS     &   \result{88.0} {        6.2 } & \result{ 44.3 }{     23.7 }&     \result{90.7} {          5.1 }&     \result{ 82.5 }{           8.9 }& \result{  27.1 }{       11.9 }\\  
&                    & TB       &   \result{91.5} {        4.9 } & \result{ 51.3 }{     25.0 }&     \result{93.6} {          3.8 }&     \result{ 87.7 }{           7.3 }& \result{  28.1 }{       12.5 }\\
&                    & TF       &   \result{89.8} {        6.1 } & \result{ 34.9 }{     23.0 }&     \result{91.6} {          5.9 }&     \result{ 85.6 }{           8.0 }& \result{  22.4 }{       11.5 }\\    
&                    & TF-ADJ   &   \result{92.3} {        2.9 } & \result{ 27.3 }{     14.8 }&     \result{94.4} {          2.0 }&     \result{ 88.8 }{           5.1 }& \result{  18.3 }{        7.2 }\\   
& $GPT$ & BAE                   &   \result{68.4} {        3.8 } & \result{ 73.0 }{      5.4 }&     \result{67.8} {          3.8 }&     \result{ 66.5 }{           4.1 }& \result{  41.4 }{        2.7 }\\ 
&                    & DWB      &   \result{85.5} {        7.0 } & \result{ 38.8 }{     17.4 }&     \result{84.4} {          7.3 }&     \result{ 86.0 }{           6.9 }& \result{  24.3 }{        8.6 }\\      
&                    & IG       &   \result{80.8} {        4.8 } & \result{ 53.5 }{     16.0 }&     \result{80.1} {          4.7 }&     \result{ 78.8 }{           6.7 }& \result{  31.5 }{        8.0 }\\  
&                    & KUL      &   \result{74.2} {        7.3 } & \result{ 65.1 }{     14.2 }&     \result{74.2} {          6.8 }&     \result{ 73.0 }{           8.3 }& \result{  36.9 }{        6.5 }\\
&                    & PRU      &   \result{67.7} {        7.4 } & \result{ 74.2 }{      9.1 }&     \result{68.2} {          7.5 }&     \result{ 66.0 }{           6.8 }& \result{  41.9 }{        4.5 }\\
&                    & PWWS     &   \result{77.9} {        8.5 } & \result{ 53.7 }{     18.6 }&     \result{76.3} {          8.8 }&     \result{ 78.0 }{           9.4 }& \result{  31.8 }{        9.3 }\\ 
&                    & TB.      &   \result{82.8} {        6.2 } & \result{ 46.7 }{     19.3 }&     \result{81.5} {          5.8 }&     \result{ 82.5 }{           8.2 }& \result{  28.2 }{        9.6 }\\  
&                    & TF       &   \result{80.2} {        7.9 } & \result{ 48.6 }{     18.5 }&     \result{78.3} {          7.9 }&     \result{ 80.6 }{           9.3 }& \result{  29.2 }{        9.2 }\\      
&                    & TF-ADJ   &   \result{69.2} {        4.7 } & \result{ 70.3 }{      9.1 }&     \result{69.8} {          4.5 }&     \result{ 67.9 }{           6.1 }& \result{  39.5 }{        4.9 }\\  
& $D_{HM}$ & BAE                &   \result{\textbf{91.3}} {        7.2 } & \result{\textbf{ 22.9 }}{     20.6 }&     \result{\textbf{91.8}} {          8.1 }&     \result{\textbf{ 89.4 }}{           7.0 }& \result{\textbf{  16.4} }{       10.3 }\\   
&                    & DWB      &   \result{\textbf{91.6}} {        3.7 } & \result{\textbf{ 48.8 }}{     15.0 }&     \result{\textbf{93.0}} {          3.6 }&     \result{\textbf{ 88.8 }}{           3.8 }& \result{\textbf{  26.8} }{        7.5 }\\  
&                    & IG       &   \result{\textbf{90.1}} {        1.3 } & \result{\textbf{ 60.9 }}{      9.1 }&     \result{\textbf{92.2}} {          1.7 }&     \result{\textbf{ 86.6 }}{           1.9 }& \result{\textbf{  32.7} }{        4.3 }\\   
&                    & KUL      &   \result{\textbf{91.9}} {        3.4 } & \result{\textbf{ 47.0 }}{     17.5 }&     \result{\textbf{93.0}} {          3.2 }&     \result{\textbf{ 90.2 }}{           4.4 }& \result{\textbf{  25.6} }{        9.0 }\\  
&                    & PRU      &   \result{\textbf{85.8}} {        5.9 } & \result{\textbf{ 68.9 }}{     12.9 }&     \result{\textbf{88.4}} {          5.4 }&     \result{\textbf{ 81.8 }}{           6.2 }& \result{\textbf{  36.7} }{        6.4 }\\  
&                    & PWWS     &   \result{\textbf{91.3}} {        3.7 } & \result{\textbf{ 28.6 }}{     15.4 }&     \result{\textbf{93.0}} {          3.5 }&     \result{\textbf{ 87.9 }}{           4.0 }& \result{\textbf{  19.3} }{        7.7 }\\ 
&                    & TB       &   \result{\textbf{94.6}} {        2.4 } & \result{\textbf{ 32.7 }}{     16.4 }&     \result{\textbf{95.7}} {          2.2 }&     \result{\textbf{ 92.3 }}{           2.9 }& \result{\textbf{  18.8} }{        8.2 }\\  
&                    & TF       &   \result{\textbf{93.3}} {        3.9 } & \result{\textbf{ 18.3 }}{     13.2 }&     \result{\textbf{94.1}} {          4.3 }&     \result{\textbf{ 91.5 }}{           3.9 }& \result{\textbf{  14.1} }{        6.6 }\\ 
&                    & TF-ADJ   &   \result{\textbf{96.9}} {        2.0 } & \result{\textbf{  7.9 }}{     10.7 }&     \result{\textbf{97.8}} {          1.2 }&     \result{\textbf{ 95.3 }}{           4.3 }& \result{\textbf{   8.4} }{        5.3 }\\  \hline
\texttt{ROB} & $D_{M}$ & BAE    &   \result{87.5} {        7.4 } & \result{ 39.4 }{     16.8 }&     \result{88.7} {          8.2 }&     \result{ 84.4 }{           7.7 }& \result{  24.6 }{        8.4 }\\ 
&                    & DWB      &   \result{90.4} {        6.0 } & \result{ 52.4 }{     23.2 }&     \result{92.4} {          4.9 }&     \result{ 86.6 }{           8.4 }& \result{  28.6 }{       11.6 }\\   
&                    & IG       &   \result{89.6} {        5.0 } & \result{ 60.8 }{     22.0 }&     \result{92.0} {          3.9 }&     \result{ 85.2 }{           7.5 }& \result{  32.7 }{       11.0 }\\   
&                    & KUL      &   \result{90.1} {        4.2 } & \result{ 57.0 }{     20.1 }&     \result{92.3} {          3.6 }&     \result{ 85.5 }{           5.8 }& \result{  30.7 }{       10.1 }\\  
&                    & PRU      &   \result{87.5} {        6.8 } & \result{ 55.3 }{     23.6 }&     \result{89.6} {          5.6 }&     \result{ 84.2 }{           9.0 }& \result{  29.8 }{       11.9 }\\ 
&                    & PWWS     &   \result{89.5} {        5.7 } & \result{ 36.0 }{     20.7 }&     \result{91.6} {          5.1 }&     \result{ 85.0 }{           7.2 }& \result{  22.9 }{       10.3 }\\ 
&                    & TB       &   \result{92.5} {        4.6 } & \result{ 44.9 }{     22.5 }&     \result{94.3} {          3.5 }&     \result{ 89.1 }{           6.9 }& \result{  24.9 }{       11.2 }\\
&                    & TF       &   \result{91.0} {        4.6 } & \result{ 29.5 }{     17.3 }&     \result{92.7} {          4.3 }&     \result{ 87.3 }{           6.1 }& \result{  19.7 }{        8.6 }\\    
&                    & TF-ADJ   &   \result{91.3} {        3.3 } & \result{ 33.2 }{     21.2 }&     \result{93.6} {          2.3 }&     \result{ 87.2 }{           5.6 }& \result{  20.5 }{        9.6 }\\   
& $GPT$ & BAE                   &   \result{69.3} {        4.9 } & \result{ 71.6 }{      6.9 }&     \result{68.0} {          4.8 }&     \result{ 67.4 }{           5.4 }& \result{  40.7 }{        3.4 }\\          
&                    & DWB      &   \result{88.4} {        5.2 } & \result{ 31.5 }{     13.7 }&     \result{87.4} {          5.6 }&     \result{ 88.8 }{           5.3 }& \result{  20.6 }{        6.8 }\\ 
&                    & IG       &   \result{82.5} {        5.5 } & \result{ 51.0 }{     16.1 }&     \result{82.3} {          5.2 }&     \result{ 80.6 }{           7.2 }& \result{  30.2 }{        8.1 }\\  
&                    & KUL      &   \result{73.6} {        8.3 } & \result{ 66.0 }{     16.1 }&     \result{73.7} {          7.0 }&     \result{ 72.6 }{           9.5 }& \result{  37.1 }{        7.2 }\\  
&                    & PRU.     &   \result{68.7} {        9.3 } & \result{ 71.3 }{     10.6 }&     \result{69.8} {          8.8 }&     \result{ 67.7 }{           7.6 }& \result{  40.0 }{        4.8 }\\ 
&                    & PWWS.    &   \result{80.0} {        7.3 } & \result{ 50.8 }{     17.6 }&     \result{78.9} {          7.1 }&     \result{ 79.4 }{           8.8 }& \result{  30.4 }{        8.8 }\\
&                    & TB       &   \result{85.8} {        5.0 } & \result{ 39.8 }{     17.3 }&     \result{84.9} {          4.4 }&     \result{ 85.4 }{           7.4 }& \result{  24.8 }{        8.6 }\\ 
&                    & TF       &   \result{81.5} {        7.5 } & \result{ 47.6 }{     18.2 }&     \result{80.0} {          7.6 }&     \result{ 81.3 }{           8.9 }& \result{  28.7 }{        9.1 }\\ 
&                    & TF-ADJ   &   \result{71.0} {        5.5 } & \result{ 72.6 }{     10.3 }&     \result{71.9} {          5.3 }&     \result{ 68.9 }{           7.4 }& \result{  40.5 }{        4.5 }\\  
& $D_{HM}$ & BAE                &   \result{\textbf{90.7}} {        7.9 } & \result{\textbf{ 22.0} }{     19.6 }&     \result{\textbf{90.3}} {         10.1 }&     \result{ \textbf{89.4 }}{           6.8 }& \result{  \textbf{15.9 }}{        9.8 }\\   
&                    & DWB      &   \result{\textbf{93.9}} {        2.4 } & \result{\textbf{ 37.4} }{     15.7 }&     \result{\textbf{94.5}} {          2.8 }&     \result{ \textbf{91.8 }}{           3.7 }& \result{  \textbf{21.1 }}{        7.9 }\\  
&                    & IG       &   \result{\textbf{92.5}} {        1.0 } & \result{\textbf{ 47.8} }{     15.4 }&     \result{\textbf{93.5}} {          1.7 }&     \result{ \textbf{89.3 }}{           2.8 }& \result{  \textbf{26.1 }}{        7.5 }\\     
&                    & KUL      &   \result{\textbf{94.2}} {        3.3 } & \result{\textbf{ 36.5} }{     20.3 }&     \result{\textbf{94.6}} {          3.6 }&     \result{ \textbf{92.7 }}{           4.4 }& \result{  \textbf{20.3 }}{       10.6 }\\   
&                    & PRU      &   \result{\textbf{89.0}} {        5.7 } & \result{\textbf{ 55.1} }{     21.1 }&     \result{\textbf{89.8}} {          6.6 }&     \result{ \textbf{86.2 }}{           7.1 }& \result{  \textbf{29.7 }}{       10.8 }\\  
&                    & PWWS     &   \result{\textbf{92.8}} {        3.3 } & \result{\textbf{ 20.5} }{     12.0 }&     \result{\textbf{93.4}} {          4.4 }&     \result{ \textbf{90.1 }}{           3.7 }& \result{  \textbf{15.2 }}{        6.0 }\\ 
&                    & TB       &   \result{\textbf{95.7}} {        2.2 } & \result{\textbf{ 26.4} }{     17.8 }&     \result{\textbf{96.4}} {          1.9 }&     \result{ \textbf{93.8 }}{           3.8 }& \result{  \textbf{15.6 }}{        8.9 }\\    
&                    & TF       &   \result{\textbf{94.3}} {        3.1 } & \result{\textbf{ 14.2} }{     10.3 }&     \result{\textbf{94.7}} {          3.9 }&     \result{ \textbf{92.5 }}{           3.7 }& \result{  \textbf{12.0 }}{        5.2 }\\    
&                    & TF-ADJ   &   \result{\textbf{96.8}} {        2.8 } & \result{\textbf{ 10.5} }{     16.5 }&     \result{\textbf{97.7}} {          1.8 }&     \result{ \textbf{94.9 }}{           4.9 }& \result{  \textbf{ 9.2 }}{        8.3 }\\  
\bottomrule  
\end{tabular}}  
\end{table*}

\subsection{Fine grained analysis per dataset}\label{ssec:analyse_per_ds_sum}
We report in \autoref{tab:all_average_supp}, the performances on {\BENCHMARK} averaged over the datasets for different detector configurations. We observe that {\DETECTOR} achieves the best results on 2 out of the 3 datasets. Overall, it is interesting to note that {\DETECTOR}'s performances are more consistent compared to Mahanalobis when changing the feature representation (\textit{i.e.,} using $f_\theta^{L}$ instead of $f_\theta^{L+1}$). 
\\\noindent\textbf{Takeaways.} This validates that the half-space mass is better for detecting textual adversarial attacks than the widely used Mahanalobis score.
\begin{table*}[!ht]
    \centering
        \caption{Average performance on {\BENCHMARK} per training dataset.}
    \label{tab:all_average_supp}
\begin{tabular}{cccccccc}  \toprule     
&                    &        &  {\AUROC} &  {\FPR} &{\AUPRIN} &{\AUPROUT}&{\ERR} \\\hline
ag-news & $D_{M}$ & $f_{\theta}^L$        &   \result{92.5} {        1.6} & \result{ 37.2 }     {14.3} &    \result{ 94.1} {          1.3} &    \result{  89.9} {           2.3} & \result{  22.2 }{        6.2} \\    
\rowcolor{LightGreen} &                    & $f_{\theta}^{L+1}$ &   \result{94.8} {        2.1} & \result{ 23.9 }     {17.3} &    \result{ 95.6} {          2.3} &    \result{  93.0} {           2.7} & \result{  15.5 }{        7.7} \\ 
& $GPT2$ & $f_{\theta}^L$                 &   \result{84.1} {        8.2} & \result{ 41.5 }     {20.3} &    \result{ 83.2} {          7.9} &    \result{  84.2} {           9.3} & \result{  25.6 }{       10.1} \\   
& {\DETECTOR}&                &   \result{92.8} {        1.9} & \result{ 39.8 }     {21.4} &    \result{ 94.8} {          1.5} &    \result{  88.7} {           2.8} & \result{  23.4 }{        9.6} \\    
&    $D_{HM}$                 & $f_{\theta}^{L+1}$ &   \result{91.9} {        2.1} & \result{ 46.1 }     {18.8} &    \result{ 93.9} {          1.7} &    \result{  88.6} {           2.9} & \result{  26.5 }{        8.3} \\ 
imdb & $D_{M}$ & $f_{\theta}^L$           &   \result{93.6} {        2.1} & \result{ 30.5 }     {17.5} &    \result{ 95.3} {          1.6} &    \result{  90.2} {           3.5} & \result{  18.9 }{        7.9} \\     
&                    & $f_{\theta}^{L+1}$ &   \result{94.2} {        3.8} & \result{ 18.6 }     {18.0} &    \result{ 93.9} {          5.4} &    \result{  92.6} {           4.0} & \result{  12.9 }{        8.2} \\ 
& $GPT2$ & $f_{\theta}^L$                 &   \result{71.7} {        8.7} & \result{ 63.8 }     {15.4} &    \result{ 70.5} {          7.7} &    \result{  72.3} {           9.8} & \result{  36.4 }{        7.4} \\    
\rowcolor{LightGreen} & {\DETECTOR} &             &   \result{96.6} {        2.3} & \result{ 14.5 }     {17.3} &    \result{ 97.3} {          1.8} &    \result{  95.2} {           3.3} & \result{  10.8 }{        7.9} \\  
&     $D_{HM}$                & $f_{\theta}^{L+1}$ &   \result{96.7} {        2.7} & \result{ 12.8 }     {15.9} &    \result{ 96.7} {          3.9} &    \result{  95.6} {           3.6} & \result{   9.9 }{        7.3} \\  
sst2 & $D_{M}$ & $f_{\theta}^L$           &   \result{82.4} {        4.7} & \result{ 71.3 }     {17.5} &    \result{ 85.2} {          5.1} &    \result{  76.3} {           5.6} & \result{  39.0 }{        8.1} \\       
&                    & $f_{\theta}^{L+1}$ &   \result{81.6} {        7.7} & \result{ 52.4 }     {20.0} &    \result{ 76.7} {         10.0} &    \result{  80.6} {           7.2} & \result{  29.6 }{        9.7} \\  
& $GPT2$ & $f_{\theta}^L$                 &   \result{74.3} {        6.3} & \result{ 67.0 }     {11.4} &    \result{ 74.7} {          6.0} &    \result{  71.3} {           6.8} & \result{  38.2 }{        5.5} \\    
\rowcolor{LightGreen} & {\DETECTOR} &            &   \result{88.7} {        5.7} & \result{ 42.8 }     {19.3} &    \result{ 88.8} {          6.0} &    \result{  87.3} {           6.4} & \result{  24.6 }{        9.5} \\  
&    $D_{HM}$                 & $f_{\theta}^{L+1}$ &   \result{88.6} {        5.9} & \result{ 42.4 }     {19.4} &    \result{ 88.0} {          6.7} &    \result{  87.3} {           6.2} & \result{  24.5 }{        9.4} \\\hline
\end{tabular}  
\end{table*}

\subsection{Extended figures for \autoref{ssec:identifying_key}}\label{ssec:analyse_per_pretrained_sum}
\begin{figure*}[!ht]
\centering
\begin{subfigure}[b]{0.8\textwidth}
 \centering
 \includegraphics[width=\textwidth]{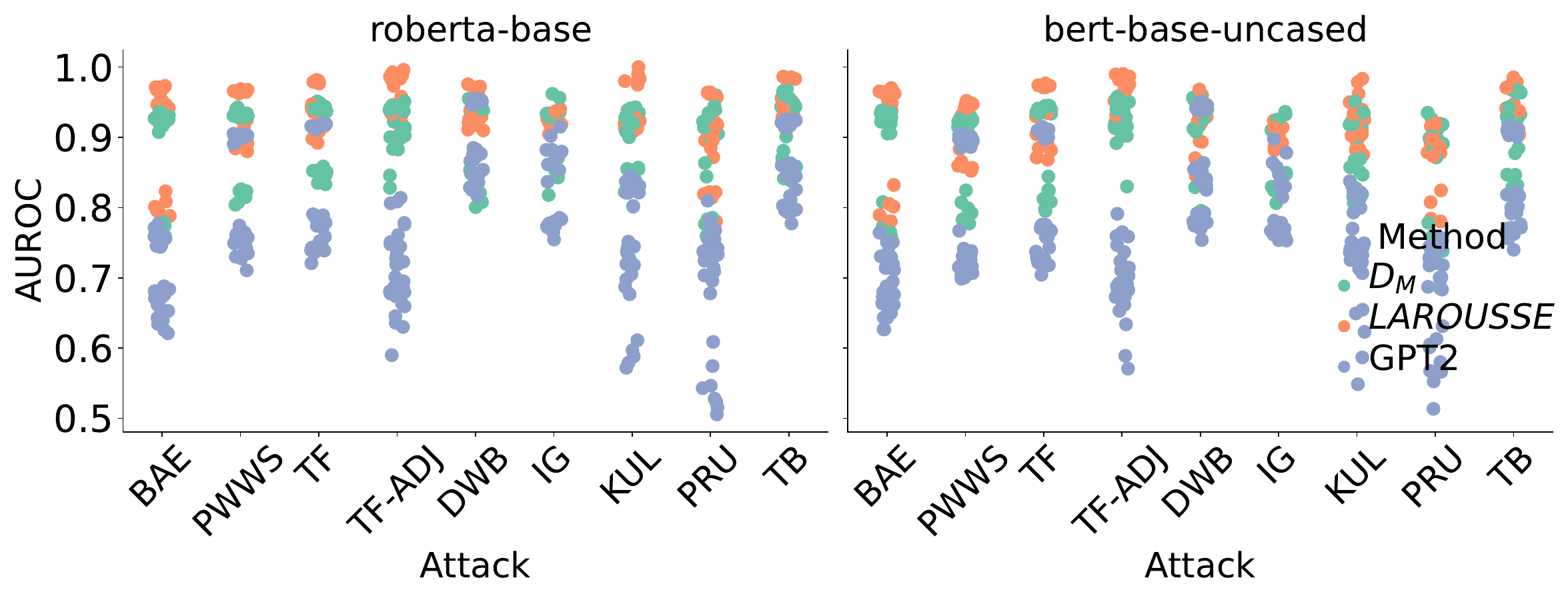}
 \caption{{\AUROC}}
 \label{fig:y equals x}
\end{subfigure}
\hfill
\begin{subfigure}[b]{0.8\textwidth}
 \centering
 \includegraphics[width=\textwidth]{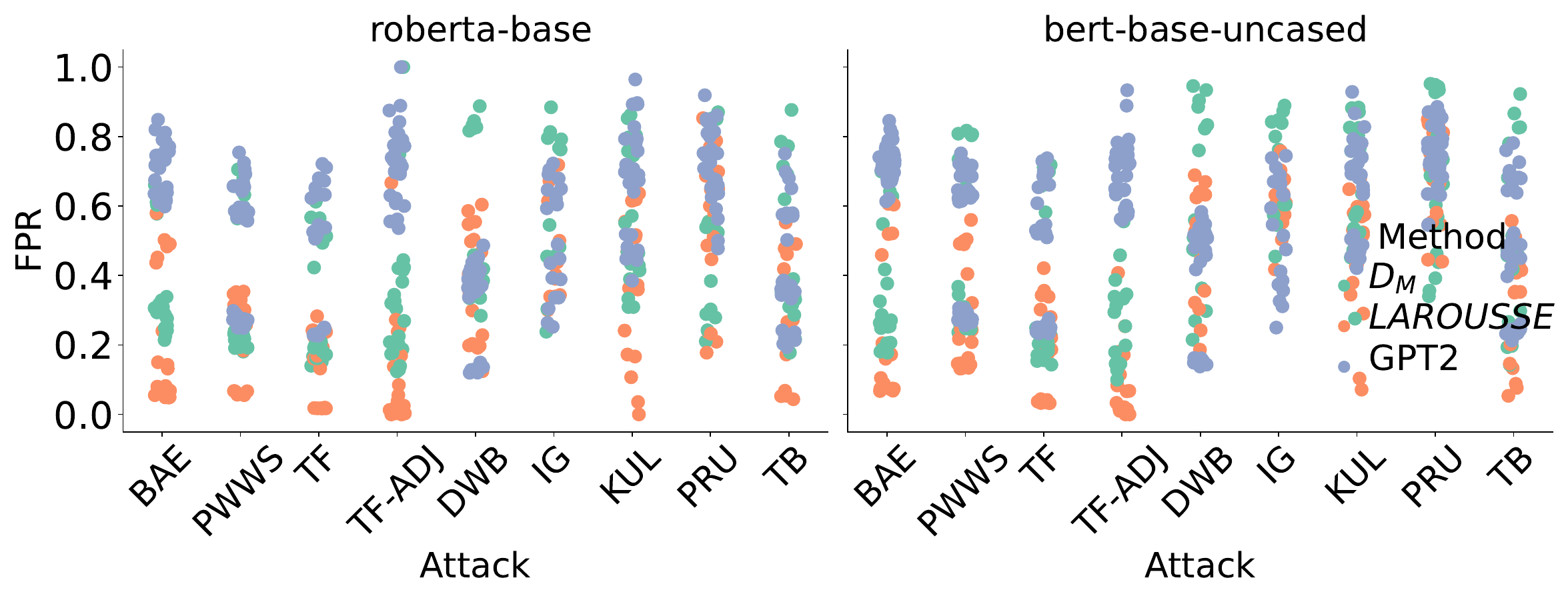}
 \caption{{\FPR}}
 \label{fig:three sin x}
\end{subfigure}
\begin{subfigure}[b]{0.8\textwidth}
 \centering
 \includegraphics[width=\textwidth]{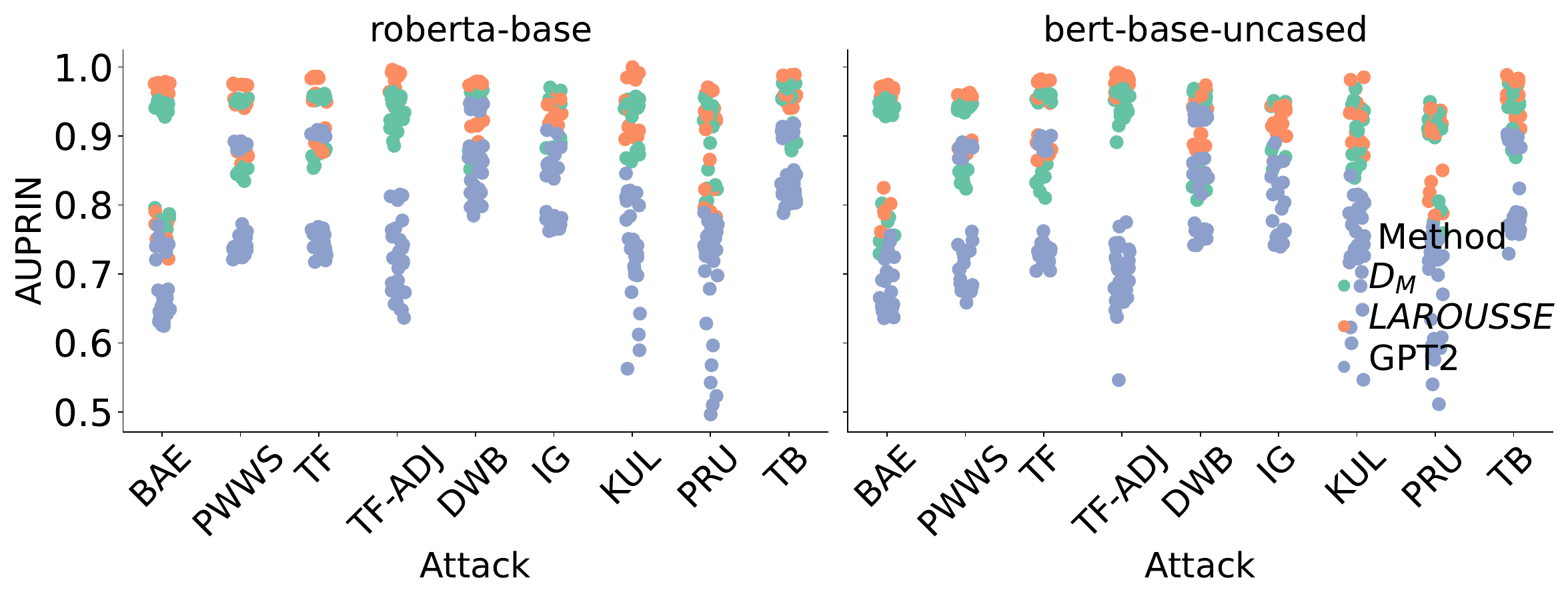}
 \caption{{\AUPRIN}}
 \label{fig:three sin x}
\end{subfigure}
\begin{subfigure}[b]{0.8\textwidth}
 \centering
 \includegraphics[width=\textwidth]{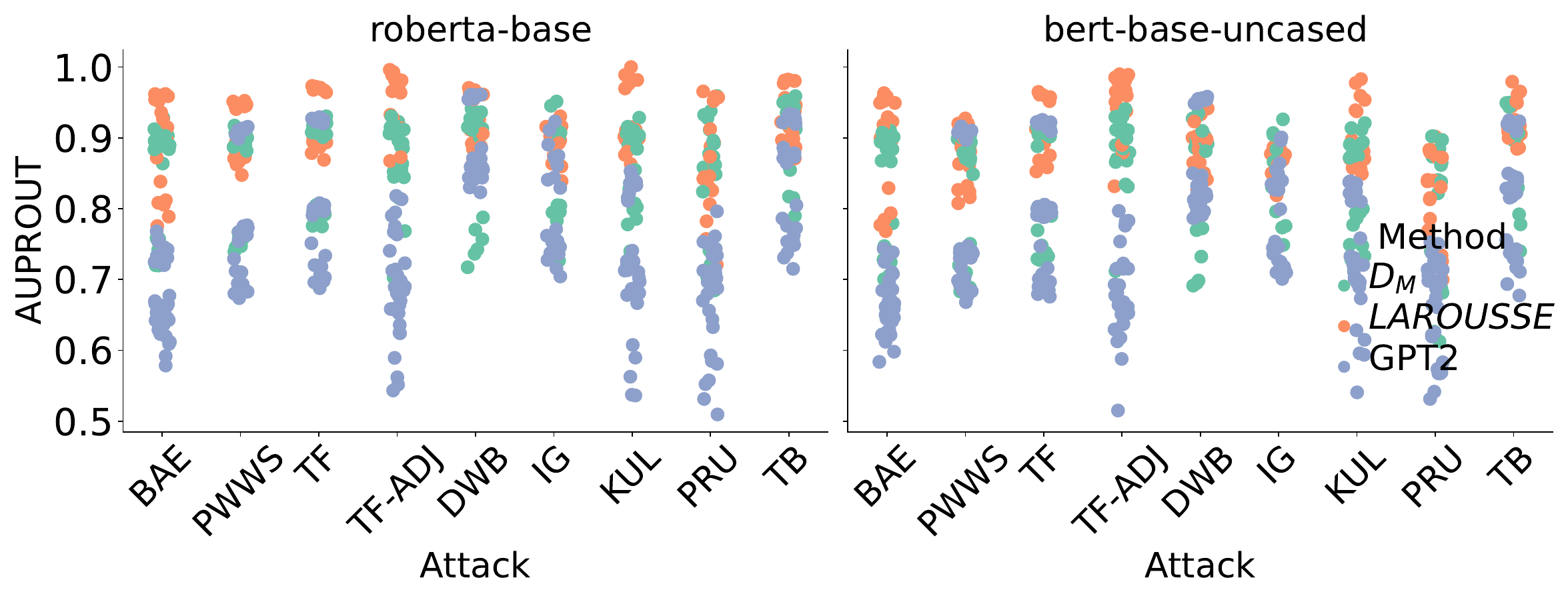}
 \caption{{\AUPROUT}}
 \label{fig:three sin x}
\end{subfigure}
\caption{Extended figures for \autoref{ssec:identifying_key}. In these figures, we report the GPT2 baseline and the performance of all the detectors in terms of {\AUROC}, {\AUPRIN},  {\AUPROUT} and {\FPR}.}
\label{fig:all_results_per_attack_appendix}
\end{figure*}

We report in \autoref{fig:all_results_per_attack_appendix} the extended figures for \autoref{ssec:identifying_key}. The baseline detector built on GPT2 is weaker than $D_M$ and {\DETECTOR}, and  consistantly achieves lower results in term of {\AUROC}, {\AUPRIN},  {\AUPROUT} and {\FPR}.

\subsection{Analysis per dataset/per attack}\label{ssec:analyse_per_pretrained_per_attack_sum}
\begin{figure*}[!ht]
\centering
\begin{subfigure}[b]{\textwidth}
 \centering
 \includegraphics[width=\textwidth]{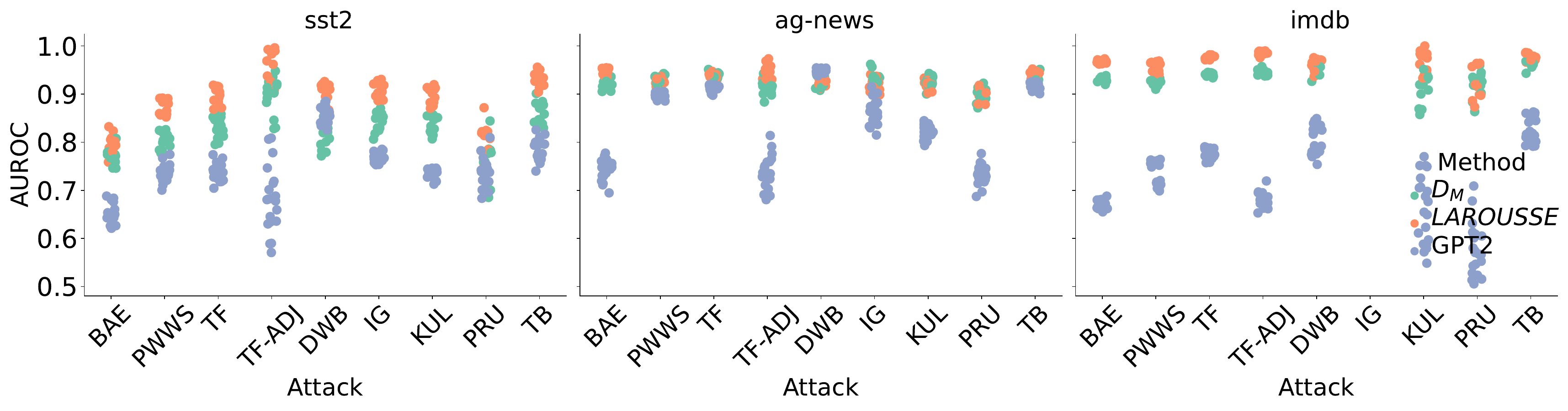}
 \caption{{\AUROC}}
 \label{fig:y equals x}
\end{subfigure}
\hfill
\begin{subfigure}[b]{\textwidth}
 \centering
 \includegraphics[width=\textwidth]{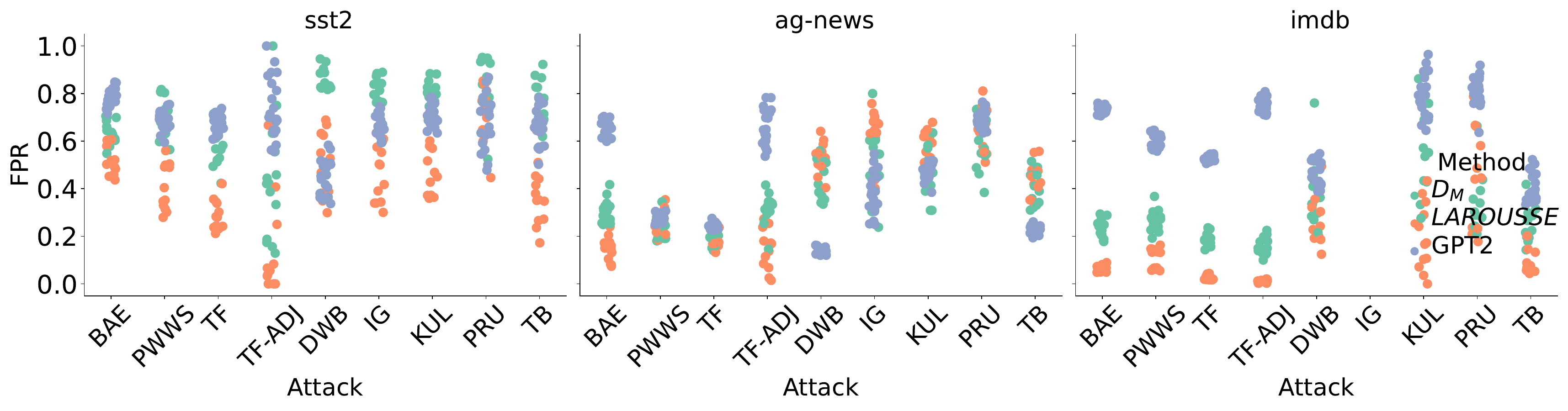}
 \caption{{\FPR}}
 \label{fig:three sin x}
\end{subfigure}
\begin{subfigure}[b]{\textwidth}
 \centering
 \includegraphics[width=\textwidth]{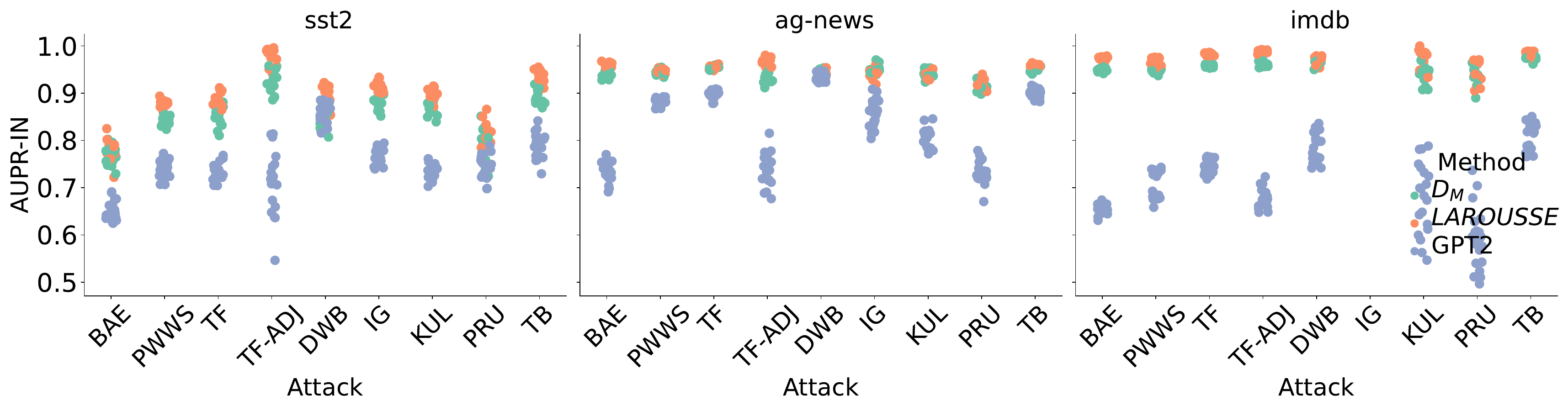}
 \caption{{\AUPRIN}}
 \label{fig:three sin x}
\end{subfigure}
\begin{subfigure}[b]{\textwidth}
 \centering
 \includegraphics[width=\textwidth]{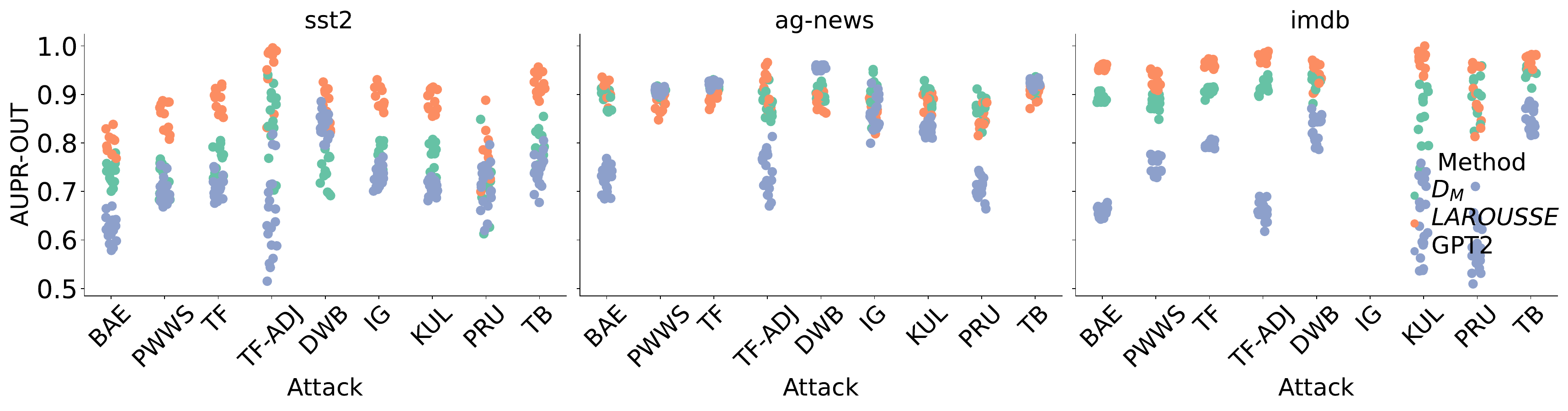}
 \caption{{\AUPROUT}}
 \label{fig:three sin x}
\end{subfigure}
\caption{Detectors performance per datasets in terms of {\AUROC}, {\AUPRIN},  {\AUPROUT} and {\FPR}.}
\label{fig:detector_performances_per_ds_supp}
\end{figure*}

We report in \autoref{fig:detector_performances_per_ds_supp} the different detectors' performances in terms of {\AUROC}, {\AUPRIN},  {\AUPROUT} and {\FPR} for the different datasets. Similar to what has been previously observed, we see a large variation in the different detectors' performance when changing both the dataset and the type of attack.

 \subsection{Comparing detection performance between semantic versus syntactic attacks}\label{ssec:syntactic}
In this section, we analyse results of the {\DETECTOR} on semantic (\textit{i.e.,} working on token) versus syntactic (\textit{i.e,} working on character) attacks. Raw and processed results are reported in \autoref{fig:semantic_analysis}.

\begin{figure*}[!ht]
\centering
\begin{subfigure}[b]{\textwidth}
 \centering
 \includegraphics[width=\textwidth]{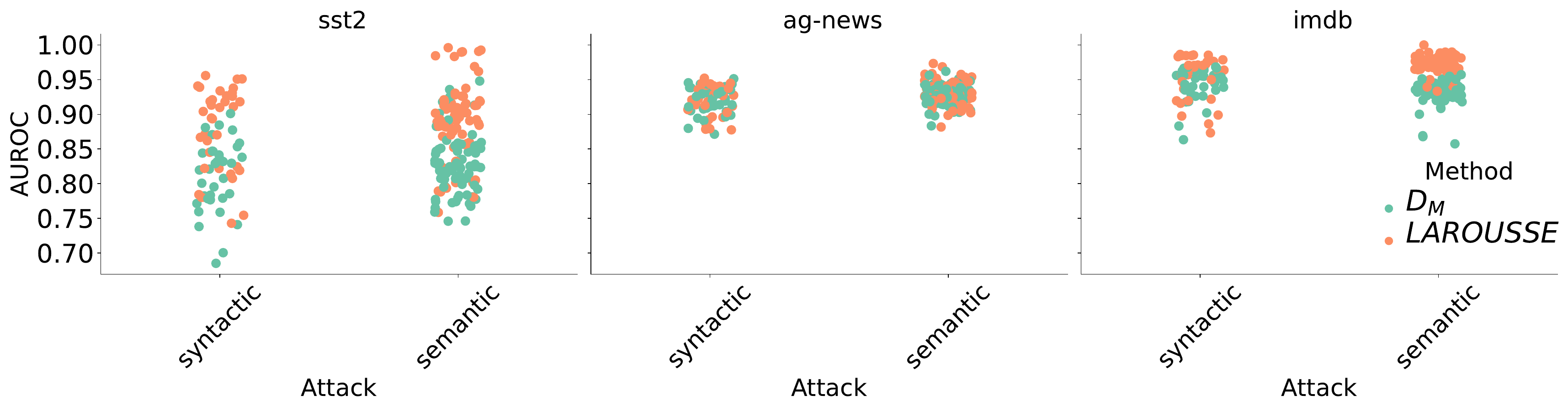}
 \caption{{\AUROC}}
 \label{fig:y equals x}
\end{subfigure}
\hfill
\begin{subfigure}[b]{\textwidth}
 \centering
 \includegraphics[width=\textwidth]{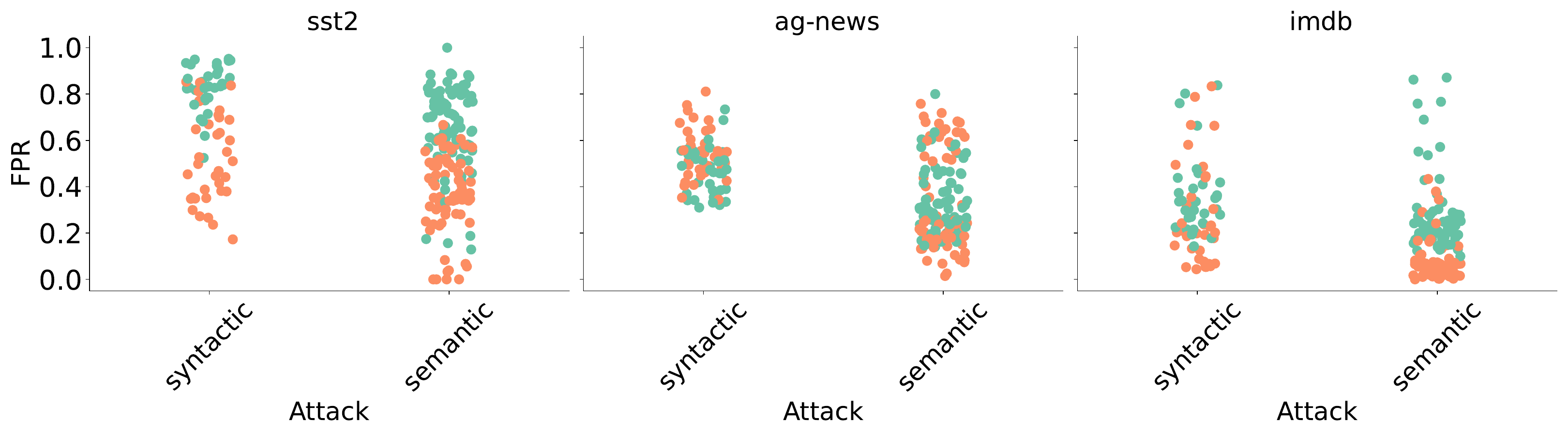}
 \caption{{\FPR}}
 \label{fig:all_fpr_sem}
\end{subfigure}
\begin{subfigure}[b]{0.4\textwidth}
 \centering
 \includegraphics[width=\textwidth]{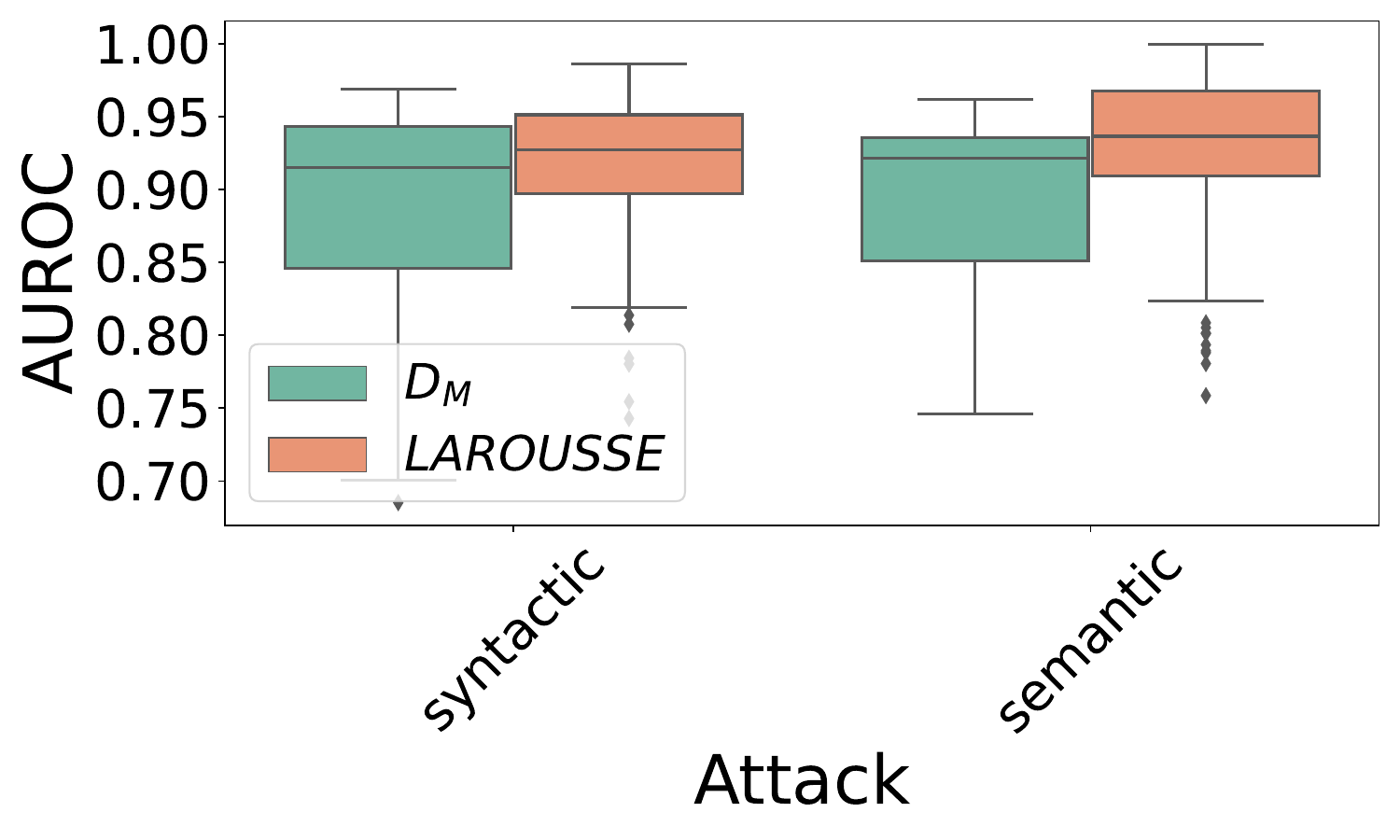}
 \caption{{\AUROC}}
 \label{fig:all_auroc_sem}
\end{subfigure}\begin{subfigure}[b]{0.4\textwidth}
 \centering
 \includegraphics[width=\textwidth]{reviewers/fpr_attack_type.pdf}
 \caption{{\FPR}}
 \label{fig:bar_fpr_sem}
\end{subfigure}
\caption{ In these figures we report the results of the semantic versus syntactic analysis in terms of {\AUROC} and {\FPR}.}
 \label{fig:bar_auroc_sem}
\end{figure*}

\textbf{Takeaways.} From \autoref{fig:all_fpr_sem}, we observe that semantic attacks are harder to detect for both our method and $D_H$.

\subsection{Towards multi-layer detectors}\label{ssec:future_rd}
A promising research direction to improve the detection methods is to develop an unsupervised strategy to combine multiple-layer representations of the pre-trained encoders \cite{gomes2022igeood,sastry2020detecting}. To the best of our knowledge, this has never been shown to be useful for text data.

\begin{figure}[!ht]
    \centering
        \includegraphics[width=0.4\textwidth]{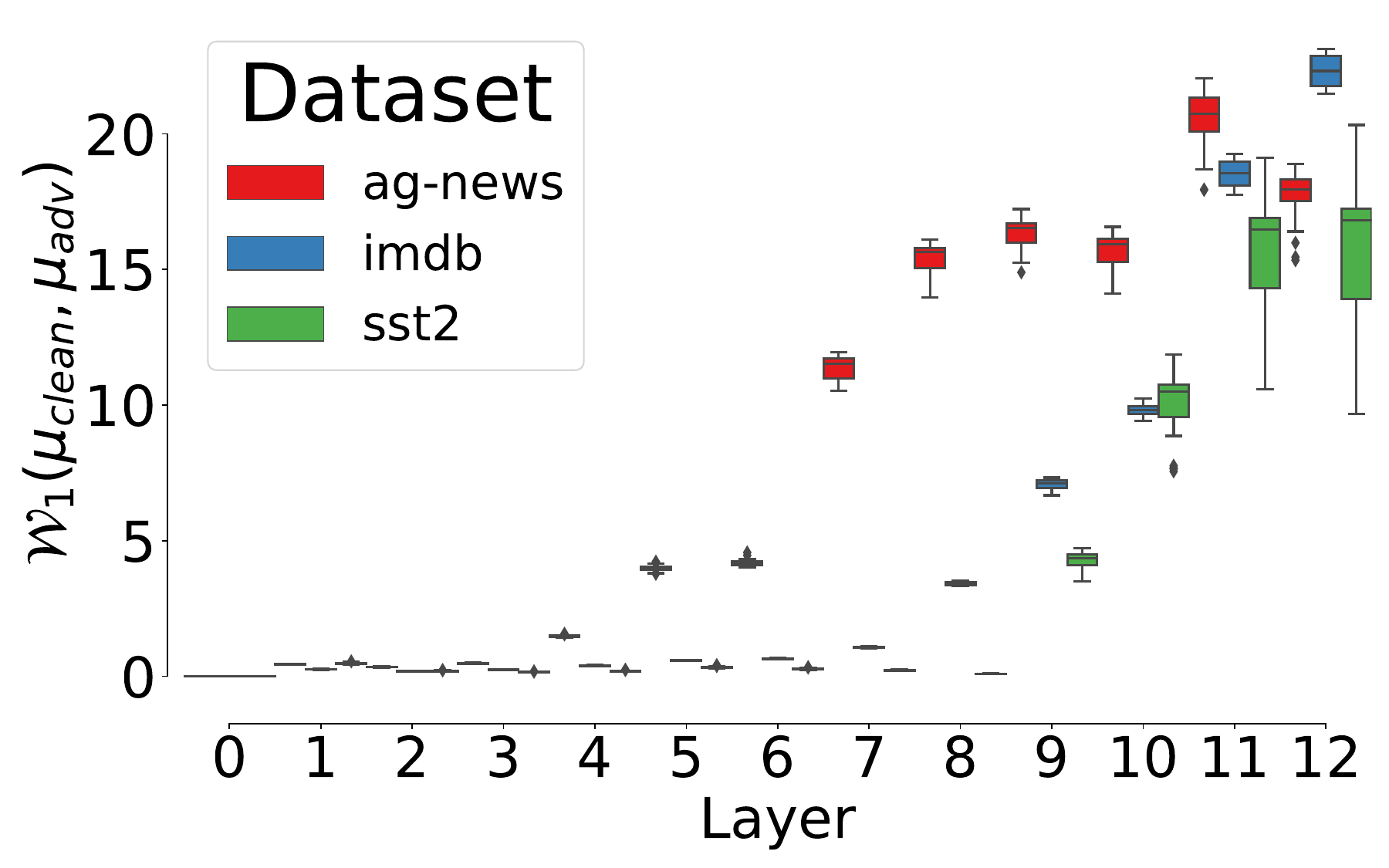}
    \caption{$\mathcal{W}_1\left(\mu_{clean},\mu_{adv}\right)$.}\label{fig:impossibility}
\end{figure}

\noindent \textbf{Setting.} In this experiment, we aim to quantify the power of each layer to discriminate between clean and adversarial samples. To measure this ability, we rely on Wasserstein distance ($\mathcal{W}_1$; see \cite{Peyre}). Given two empirical distributions, $\mathcal{W}_1$ finds the best possible transfer between them while minimizing the transportation cost defined by the Euclidean distance.
\autoref{fig:impossibility} reports the transportation cost ($\mathcal{W}_1$) between the empirical distributions of clean samples ($\mu_{clean}$) and adversarial samples ($\mu_{adv}$) obtained at each layer.
\\\noindent\textbf{Analysis.} The last layers of the encoder have a better ability to discriminate the adversarial samples from the clean one than the first layers. Similarly to what can be observed in \autoref{fig:all_tradeoff_main}, we observe that \texttt{IMDB} is the easiest dataset as the last encoder layer can better distinguish the adversarial samples and the clean one. Interestingly, we observe that the best layer depends on the dataset, which is consistent with the observation in NLG evaluation \cite{bert-score}, where the optimal layer is found using a validation set. Overall, the information present at the last encoder layers suggests that designing multi-layer detectors is a promising research direction.  

\subsection{Attacking our detectors}
Adversarial attack detection methods have been extensively studied in the computer vision community \cite{feinman2017detecting,ma2018characterizing,kherchouche2020natural,aldahdooh2022adversarial,aldahdooh2022revisiting} and recently a line of work on adaptive attacks \cite{carlini2017adversarial,athalye2018obfuscated,tramer2020adaptive} have emerged. {\DETECTOR} is not differentiable adding an extra layer of security: it prevents the malicious adversaries to leverage gradient computations, contrary to studied baselines (e.g Mahalanobis, GPT). Attacking {\DETECTOR} is thus a quite challenging research question that falls outside of the scope of the paper and is left as future work.

\section{Future work}
In the future, we would like to extend our adversarial detection setting to natural language generation tasks on seq2seq models
\cite{pichler2022differential,colombo-etal-2019-affect,colombo-etal-2021-novel,colombo-etal-2021-improving} and classification tasks \cite{chapuis-etal-2020-hierarchical,colombo-etal-2022-learning,colombo2022best,himmi2023towards} as well as NLG evaluation \cite{colombo-etal-2021-automatic,colombo2022glass,colombo-etal-2021-beam,colombo-etal-2021-code,colombo-etal-2021-improving} and Safe AI \cite{colombo2022beyond,picot2022adversarial,picot2022simple,darrin2022rainproof,darrin2023unsupervised}.
\section{Computation time comparison between HM and Mahalanobis depths}

\begin{figure*}
    \centering
    \begin{tabular}{cc}
       \includegraphics[scale=0.4, trim=2.5cm 0 0 0]{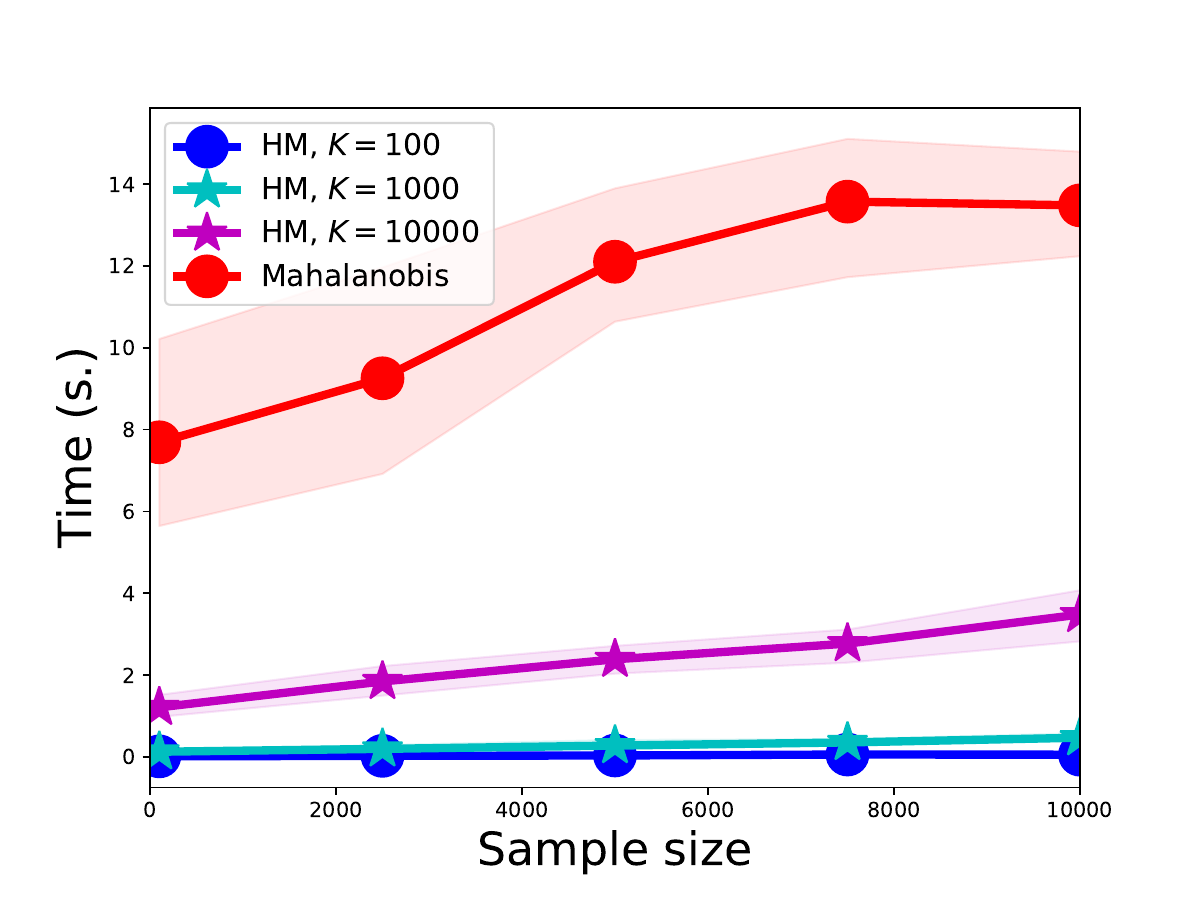}  & \includegraphics[scale=0.4, trim=2.5cm 0 0 0]{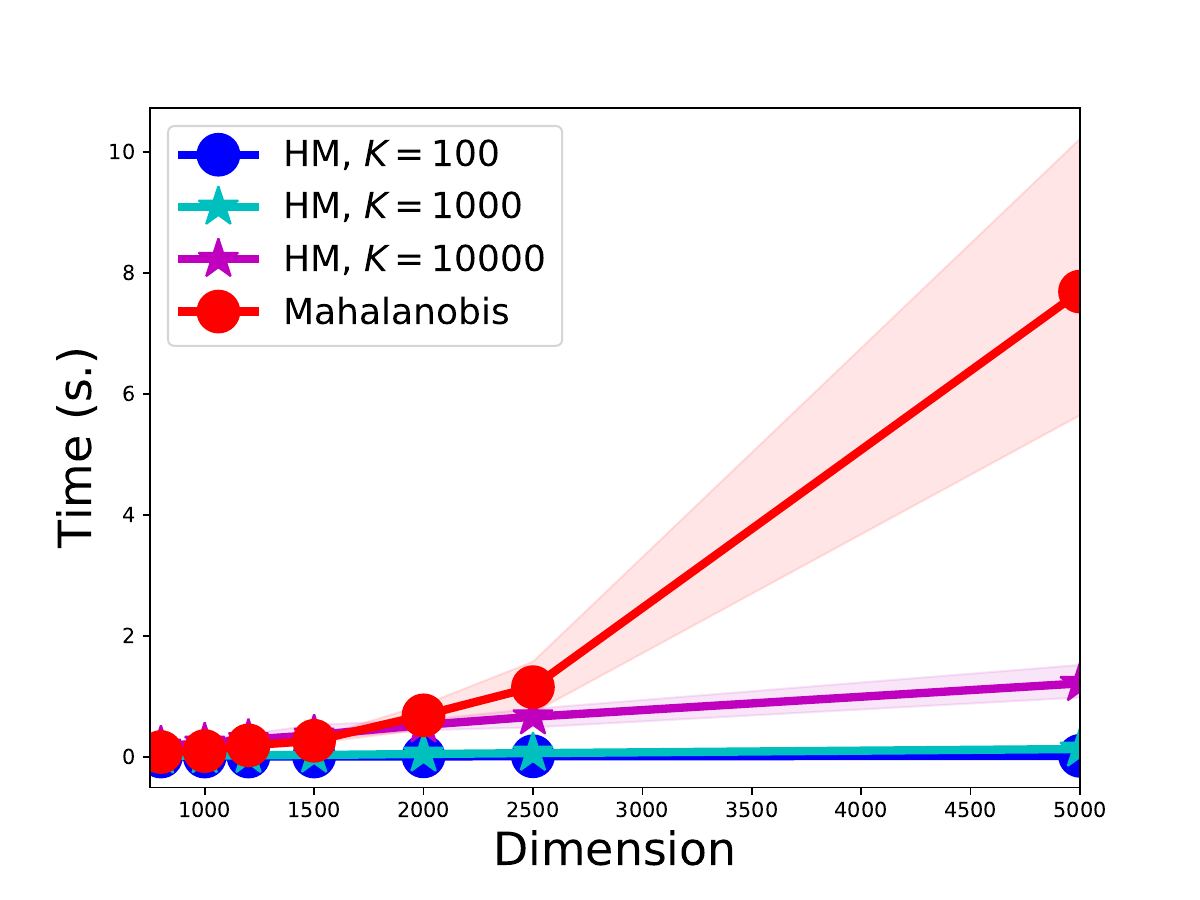}
    \end{tabular}
    \caption{Computation time of HM ($K \in \{100, 1000, 10000\}$) and Mahalanobis depths for various sample sizes (left) and dimensions (right).}
    \label{fig:computation_time}
\end{figure*}

In this part, we compare the computation time between the HM and the Mahalanobis depths. Precisely, we want to compare the time of computing these data depths of an element $x\in \mathbb{R}^d$  w.r.t. a dataset. This experiment is conducted as follows: several datasets (varying dimension $\{800, 1000, 1200, 1500, 2000, 2500, 5000\}$ and sample size $\{100, 2500, 5000, 7500, 10000\}$) are sampled from centered Gaussian distributions $\mathcal{N}(0, \Sigma)$ where $\Sigma$ follows a Wishart distribution. Therefore, we compute the depth of $\mathbf{0}$ w.r.t. each of these datasets. This procedure is repeated 10 times. We report their mean computation time as well as 10-90\% quantiles in Figure~\ref{fig:computation_time} highlighting the computational benefits of using the HM depth over the Mahalanobis distance. The dimension is 5000 on the left picture while the sample size is fixed to 100 on the right picture.

\end{document}